\documentclass[twoside,11pt]{article}

\usepackage[preprint]{jmlr2e}

\usepackage{amsmath}
\usepackage{amsfonts}
\usepackage{mathrsfs}
\usepackage{comment}
\usepackage{enumitem}
\usepackage{microtype}
\usepackage{algorithm}
\usepackage{algorithmic}
\usepackage{tabularx}
\usepackage{hhline}
\usepackage[parfill]{parskip}
\usepackage{thmtools}
\usepackage{thm-restate}
\usepackage{cleveref}

\def\environment{\mathcal{E}}

\def\actions{\mathcal{A}}
\def\policies{\mathcal{P}}
\def\observations{\mathcal{O}}
\def\histories{\mathcal{H}}

\def\states{\mathcal{S}}

\def\regret{\mathrm{Regret}}
\def\emptyH{H_0}

\def\falea{f}

\def\trans{\mathbf{P}}
\def\rbar{\overline{r}}
\def\E{\mathbb{E}}

\def\F{\mathcal{F}}

\def\Pr{\mathbb{P}}
\def\Re{\mathbb{R}}
\def\1{\mathbf{1}}

\def\nn{\nonumber}
\def\noindentproof{\noindent{\bf Proof.}$\ $}
\DeclareMathOperator*{\argmax}{arg\,max}
\DeclareMathOperator*{\argmin}{arg\,min}

\usepackage{color}
\definecolor{darkred}{rgb}{0.7,0,0}
\definecolor{darkgreen}{rgb}{0.0,0.5,0.0}
\definecolor{darkblue}{rgb}{0.0,0.0,0.5}
\definecolor{teal}{rgb}{0.0,0.5,0.5}
\newcount\Comments
\newcommand{\kibitz}[2]{\ifnum\Comments=1{\textcolor{#1}{\textrm{\footnotesize #2}}}\fi}

\renewenvironment{equation*}
{\begin{eqnarray}}{\end{eqnarray}\ignorespacesafterend}
\newcommand{\qedsymbol}{\hfill$\blacksquare$}
\newenvironment{packed_enum}{
\begin{enumerate}
  \setlength{\itemsep}{1pt}
  \setlength{\parskip}{0pt}
  \setlength{\parsep}{0pt}
}{\end{enumerate}}

\ShortHeadings{Simple Agent, Complex Environment}{Dong, Van Roy and Zhou}
\firstpageno{1}

\begin{document}

\title{Simple Agent, Complex Environment:\\ Efficient Reinforcement Learning with Agent States}

\author{\name Shi Dong \email sdong15@stanford.edu\\
       \name Benjamin Van Roy \email bvr@stanford.edu
       \\
       \addr 
       Stanford University
       \AND
       \name Zhengyuan Zhou \email zzhou@stern.nyu.edu \\
       \addr New York University}

\editor{}

\maketitle

\begin{abstract}
We design a simple reinforcement learning (RL) agent that implements an optimistic version of $Q$-learning and establish through regret analysis that this agent can operate with some level of competence in {\it any} environment.
While we leverage concepts from the literature on provably efficient RL, we consider a {\it general} agent-environment interface and provide a novel agent design and analysis.  This level of generality positions our results to inform the design of future agents for operation in complex real environments.
We establish that, as time progresses, our agent performs competitively relative to policies that require longer times to evaluate.  The time it takes to approach asymptotic performance is polynomial in the complexity of the agent's state representation and the time required to evaluate the best policy that the agent can represent.  Notably, there is no dependence on the complexity of the environment. 
The ultimate per-period performance loss of the agent is bounded by a constant multiple of a measure of distortion introduced by the agent’s state representation.  This work is the first to establish that an algorithm approaches this asymptotic condition within a tractable time frame.
\end{abstract}


\section{Introduction}
Reinforcement learning agents have demonstrated remarkable success in simulated environments.  For example, the recently developed MuZero agent \citep{schrittwieser2020muzero} learns to interact effectively with any of a broad range of environment simulators and delivers superhuman performance in playing chess, go, shogi, and arcade games.  Continuing innovations in this area aim to produce agents that can engage with increasingly complex environments -- ultimately, environments like the physical world or the World Wide Web -- which pose far greater complexity than the agent can represent.

There is a growing mathematical literature that focuses on establishing efficiency guarantees, typically in terms of sample complexity or regret bounds (\cite{kearns2002near,jaksch2010near} represent early instances).  Indeed, efficiency remains an impediment to carrying the success of reinforcement learning from simulated to real environments, in which agents must learn within reasonable time frames.  As such, the mathematical literature ought to inform future agent designs.  However, work in this area has tended to focus on restrictive classes of environments, and further, to produce bounds that depend on the number of environment states, which is effectively infinite in a complex environment.

In this paper, we aim to bridge the divide.  In particular, we extend ideas from the mathematical literature while relaxing common restrictions.  In doing so, we establish results that offer insight into how a simple agent can operate effectively in an arbitrarily complex environment.  This work contributes to multiple fronts: problem formulation, framing of learning objectives, agent design, and performance analysis.

\begin{figure}[H]
    \centering
    \includegraphics[scale=0.4]{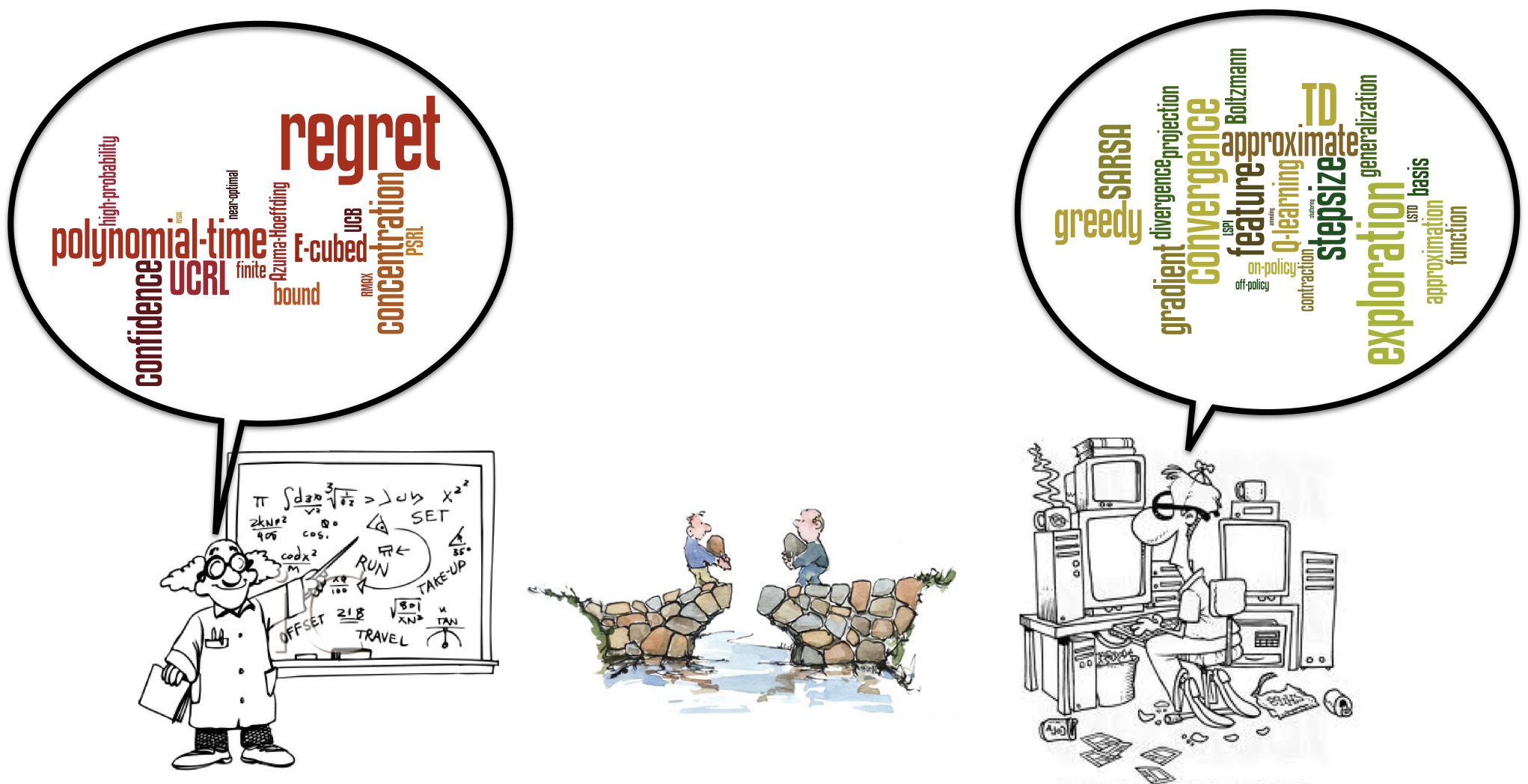}
    \caption{Bridging the divide: ``provably efficient'' reinforcement learning versus ``practical'' agent design.}
    \label{fig:bridge}
\end{figure}

\subsection{Complex Environments}
\label{se:complex-environments}

We consider an interface, as illustrated in Figure \ref{fig:interface}, which is defined by a finite action set $\actions$ and a finite observation set $\observations$.  The agent interacts with the environment by executing at each time $t$ an action $A_t$ and then registering an observation $O_{t+1}$, generating a single stream of experience $(A_0, O_1, A_1, O_2, \ldots)$.  At each time $t$, the agent selects $A_t$ based on its history $H_t = (A_0,O_1,\ldots,A_{t-1},O_t)$.  The initial history $H_0 = ()$ is empty.

The interface we have described is very general.  An agent can engage in this manner with arbitrarily complex environments.  As an example, consider an agent that interacts with the World Wide Web via a computer terminal.  Each action could encode a keystroke or mouse click or movement, while observations could take the form of pixels rendered by a monitor.  In such a context, the environment would likely be far more complex than the agent.

Environment dynamics are characterized by a function $\rho$, which assigns a probability $\rho(o|H_t, A_t) = \Pr(O_{t+1} = o |\environment, H_t, A_t)$ to each observation $o \in \observations$.  Hence, an environment is specified by a tuple $\environment = (\actions, \observations, \rho)$, with fixed sets $\actions$ and $\observations$ and an observation probability function $\rho$.  In order to accommodate complex real environments, our formulation relaxes several restrictive assumptions commonly made in the literature:

\begin{figure}[H]
    \centering
    \includegraphics[scale=0.3]{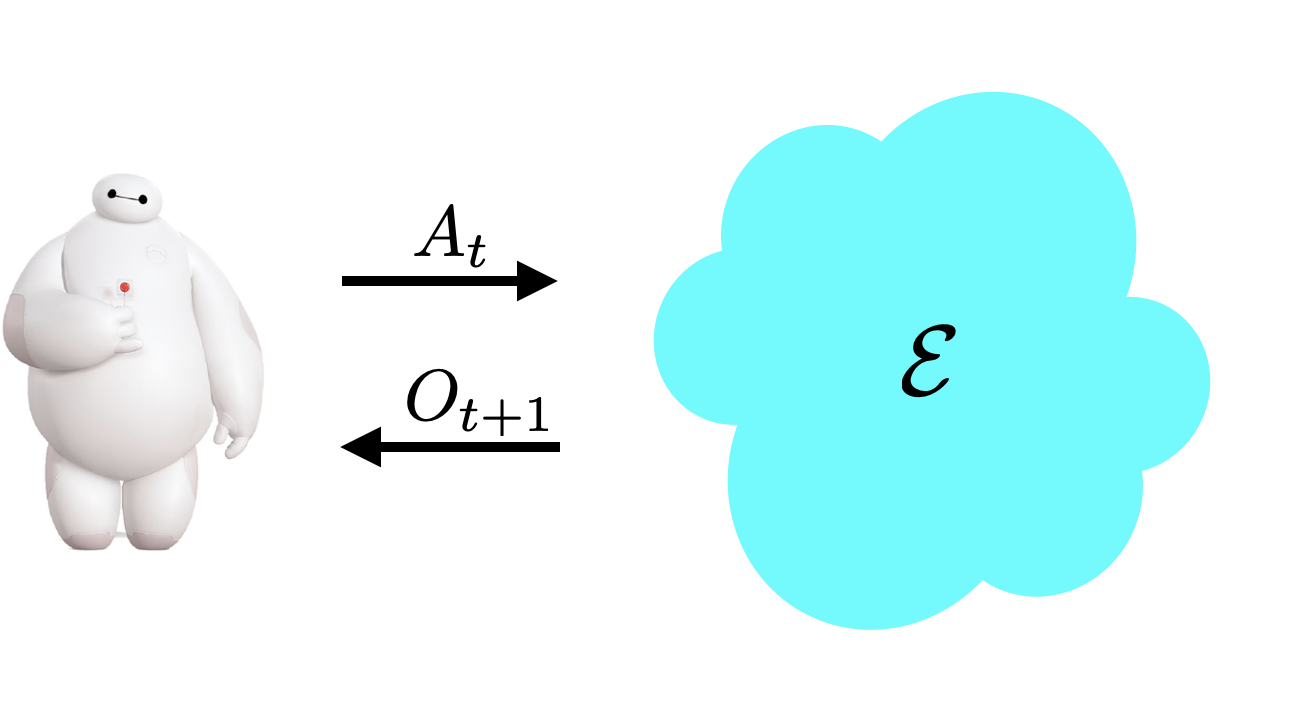}
    \caption{The agent-environment interface.}
    \label{fig:interface}
\end{figure}

\begin{enumerate}
\item We do not assume the environment is a Markov decision process (MDP), which would require observation probabilities to be independent of history conditioned on the most recent observation and action.
\item We do not assume that the environment exhibits episodic behavior, which would require that the environment occasionally ``renews.''
\item We do not assume that the performance of an optimal policy can be accurately estimated within a manageable time frame.  In a complex environment, the required time can be intractably large or even infinite.
\item We do not require that the agent be supplied with the duration $T$ of operation as input.   We consider instead a single endless stream of experience, calling for agents to perform well over any long horizon.
\end{enumerate}

\subsection{Policies and Performance}

A {\it policy} $\pi$ is a mapping from histories to action probabilities, with the probability assigned to action $a$ at history $h$ denoted by $\pi(a|h)$.  Let $\policies$ denote the set of all policies.  We denote by $\pi_{\rm agent} \in \policies$ the policy executed by the agent.  Agent design amounts to specifying this policy, typically in terms of an algorithm that samples each action $A_t$ according to $\pi_{\rm agent}(\cdot|H_t)$.

The designer's preferences are expressed in terms of a reward function $r$.  For each history $H_t$, action $A_t$, and observation $O_{t+1}$, this function prescribes a reward $R_{t+1} = r(H_t, A_t, O_{t+1})$.  We will characterize the performance of a policy in terms of expected rewards.  To formalize this notion, we build on a general probabilistic framework, the details of which are presented in Appendix \ref{se:probability}.  In this framework, actions $A_t$ and observations $O_{t+1}$ are random variables.  The observation probability function $\rho$ is also a random variable, as it is unknown to the agent designer, and consequently, the environment $\environment$ is a random variable.  As formally defined in the appendix, we use a subscript to indicate that a probability or an expectation is evaluated with actions selected by a particular policy.  For example, action probabilities satisfy $\Pr_\pi(A_t = a| H_t) = \pi(A_t = a|H_t)$, and the expected return over $T$ timesteps under policy $\pi$ is written as $\E_\pi[\sum_{t=0}^{T-1} R_{t+1}]$.  When expressing probabilities and expectations under $\pi_{\rm agent}$, we suppress subscripts. For example, $\Pr(A_t = a| H_t) = \Pr_{\pi_{\rm agent}}(A_t = a| H_t) = \pi_{\rm agent}(a|H_t)$ and $\E[\sum_{t=0}^{T-1} R_{t+1}] = \E_{\pi_{\rm agent}}[\sum_{t=0}^{T-1} R_{t+1}]$.
With this notation, we denote the average reward of a policy $\pi \in \policies$ by
\[
\lambda_\pi = \liminf_{T\to\infty} \E_\pi\left[ \frac{1}{T}\sum_{t=0}^{T-1} R_{t+1} \Big| \environment\right],
\]
and the optimal average reward by $\lambda_* = \sup_{\pi \in \policies} \lambda_\pi$.  We quantify the agent's performance relative to a {\it reference policy} $\pi\in\policies$ over $T$ timesteps in an environment $\environment$ in terms of regret:
\[
    \regret_{\pi}(T) = \E\left[\sum_{t=0}^{T-1} \big( \lambda_\pi - R_{t+1}\big) \Big|\environment\right].
\]
We will also consider a notion of regret relative to a {\it reference policy class} $\mathcal{P}' \subseteq \mathcal{P}$:
\[
\regret_{\mathcal{P}'}(T) = \sup_{\pi\in \mathcal{P}'}\regret_{\pi}(T).
\]
Note that $\regret_{\policies}(T)$ is simply the regret relative to the optimal average reward $\lambda_*$, and if there exists an optimal policy $\pi_*$ then $\regret_{\pi_*}(T) = \regret_{\policies}(T)$.  Note that the expressions defining regret are random variables, as they depend on the environment $\environment$.  From the perspective of an agent designer, a reasonable goal would be to attain low expected regret $\E[\regret_{\mathcal{P}}(T)]$ for all long durations $T$.  However, in this paper, rather than aim for optimal design, we will study the performance of fixed agents, and our results bound regret rather than expected regret.

Our regret bounds necessarily depend on the time required to assess policies.  In a complex environment, the time required to assess optimal or near-optimal policies can be arbitrarily large or even infinite.  As such, we develop bounds that depend instead on the time required to assess policies in reference classes.  In particular, our bounds indicate that, as time progresses and the agent accumulates experience, it can perform well relative to policies that take longer to assess.

\subsection{A Simple Agent}

A practical agent must operate with bounded memory and per-timestep computation.  With these constraints, the agent cannot retain and repeatedly process an ever-growing history.  Rather, the agent maintains only an {\rm agent state} $X_t$ that suffices to produce its actions.  Since $X_t$ represents all the agent retains from history, it must be updated incrementally, according to
$$X_{t+1} = f_{\rm agent}(X_t, A_t, O_{t+1}, U_{t+1}),$$
for some agent state update function $f_{\rm agent}$, where $U_{t+1}$ represents algorithmic randomness.
As discussed in \cite{lu2021reinforcement}, in popular agent designs (e.g., DQN \citep{mnih2015human}, MuZero \citep{schrittwieser2020muzero}, MPO \citep{abdolmaleki2018maximum,Song2020V-MPO:}), the agent state can be partitioned into three components:
$$\textbf{agent state } X_t = \Big(\textbf{aleatoric state } S_t, \textbf{epistemic state } P_t, \textbf{algorithmic state } Z_t\Big).$$
The aleatoric state is meant to capture salient information about the agent's current situation in the environment.  The epistemic state retains the agent's knowledge about the environment.  The algorithmic state can record information unrelated to the environment, such as readings from the agent's internal clock or internally generated random numbers.  Rewards computed by such an agent depend on history through the aleatoric state.  Letting $\states$ denote the set of aleatoric states, the reward function takes the form $r:\states\times\actions\times\observations\rightarrow\mathbb{R}$, generating rewards according to $R_{t+1} = r(S_t, A_t, O_{t+1})$.

\begin{figure}[H]
\centering
\includegraphics[scale=0.3]{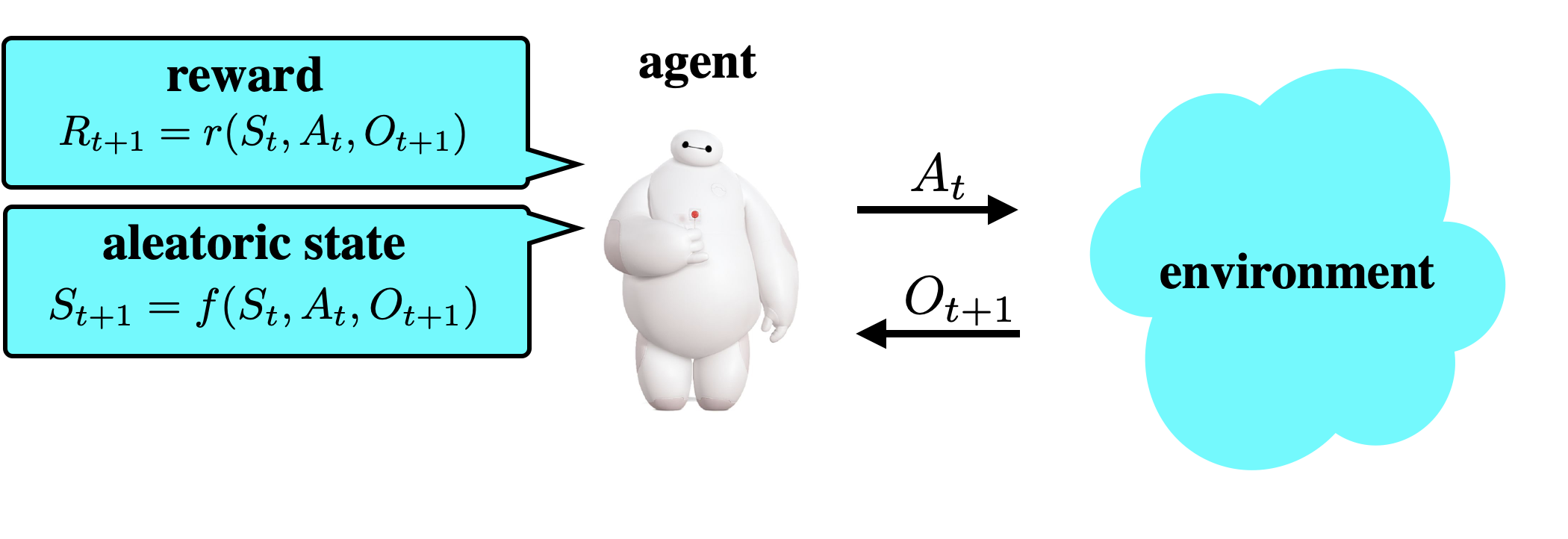}
\caption{Our agent maintains an aleatoric state and uses that to compute rewards.}
\label{fig:aleatoric-state}
\end{figure}

In this paper, we design and analyze a simple agent, which can engage with any environment after being instantiated with the following inputs:
\begin{packed_enum}
\item an initial aleatoric state $S_0 \in \states$ and update function $f:\states\times\actions\times\observations\mapsto\states$,
\item a reward function $r:\states\times\actions\times\observations\mapsto[0,1]$.
\end{packed_enum}
While we will provide a precise specification later in the paper, here we offer a rough description of how the agent operates.  Our agent updates its aleatoric state according to $S_{t+1} = f(S_t, A_t, O_{t+1})$ and uses this to compute rewards, as illustrated in Figure \ref{fig:aleatoric-state}.  The aleatoric state dynamics need not be Markovian; in particular, we can have $\Pr(S_{t+1} = s | \environment, S_t, A_t) \neq \Pr(S_{t+1} = s | \environment, H_t, A_t)$.  Our agent's epistemic state $P_t = (Q_t, N_t)$ is comprised of an action value function $Q_t:\states\times\actions\rightarrow \mathbb{R}$ and a count function $N_t:\states\times\actions\rightarrow \mathbb{Z}_+$.  Our agent's algorithmic state includes only the current time $t$.  Action values $Q_t$ are updated via an optimistic discounted Q-learning algorithm, with the discount factor and degree of optimism increasing over time.  The agent updates $N_t$ to track visitation counts, which are used to determine a suitable degree of optimism.  Each action $A_t$ is sampled uniformly from the set of greedy actions $\argmax_{a \in \actions} Q_t(S_t,a)$.  Hence, at any time, the agent can be seen as executing a policy $\pi_t(\cdot|H_t)$ for which action probabilities depend on the history $H_t$ only through the aleatoric state $S_t$.

It is worth emphasizing that our agent is not designed to offer state-of-the-art performance in simulated or real environments.  Rather, our motivation is to design an agent that is amenable to theoretical analysis, with an aim to generate insights that inform the design of future state-of-the-art agents.

\subsection{Example: Service Rate Control}
\label{se:service-station}

\begin{figure}[H]
    \centering
    \includegraphics[scale=0.15]{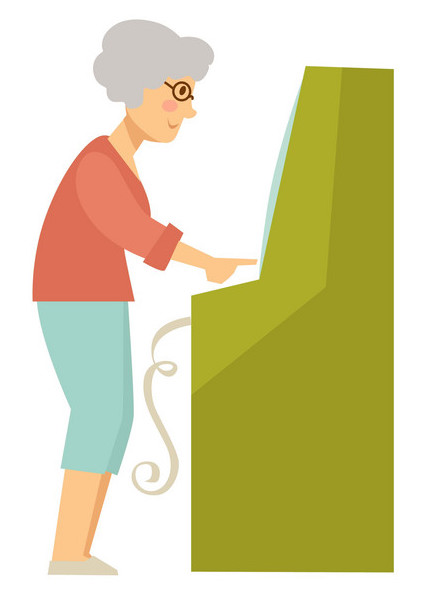}
    \caption{A service station serving a customer.}
    \label{fig:service-station}
\end{figure}

Let us consider a didactic example that, while exceedingly simple, {\it serves} to elucidate our notation and framework.  The example involves an agent operating a service station, as illustrated in Figure \ref{fig:service-station}.  At each time, there can be at most one customer present, and the agent applies a service mode -- {\it fast} or {\it slow}.  Each customer pays $\$1$ upon arrival.  No cost is incurred when the slow mode is applied or when there is no customer being served.  The fast mode of service incurs a cost of $\$0.50$ per timestep.  To maximize average reward, an agent must make choices that balance revenue against the cost of service.

This problem is one of service rate control, as studied in operations research (see, e.g., \citep{weber1987optimal,stidham1989monotonic,jo1989lagrangian,sennott2009stochastic}).  However, such work has tended to focus on agents that are effective when applied to particular stylized models that govern arrival and service rates.  Our approach instead adapts to {\it any} statistical structure, and as such, does not suffer from misspecification.  Work on reinforcement learning for control of queueing systems \citep{moallemi2008approximate,raeis2021queue} shares this spirit.  We should note that it is only in order to convey ideas in a simple and transparent manner that we focus on such a simple service system: our agent can be applied to much more complex environments, for example, involving multiple servers and queues.  

\subsubsection{Agent-Environment Interface}

From the agent's perspective, the service station can be viewed as an environment $\environment = (\actions, \observations, \rho)$.  Actions $\mathcal{A} = \{\mathrm{fast}, \mathrm{slow}\}$ identify service modes and observations $\observations = \{\mathrm{arrival}, \neg\mathrm{arrival}\} \times \{\mathrm{departure}, \neg\mathrm{departure}\}$ indicate arrivals and departures.  Hence, $A_t$ is the service mode applied over timestep $t$ and $O_{t+1}$ indicates any arrival or departure occurring by the end of the timestep.  The function $\rho$ specifies observation probabilities conditioned on history, which are initially unknown.  For example, the designer may be uncertain about customer arrival rates and how they depend on history.

\subsubsection{Aleatoric State Dynamics}

We consider an aleatoric state $S_t \in \states = \{0,1\}$ that simply indicates presence of a customer.  Since observations record arrivals and departures, there is a function $f$ for which $S_{t+1} = f(S_t, A_t, O_{t+1})$.  The service is initially vacant, so $S_0 = 0$.  Profit can be written as $R_{t+1} = r(S_t, A_t, O_{t+1})$ for some function $r$.  Of special interest are policies that select actions based only on aleatoric state; that is, the  set of policies for which $\Pr_\pi(A_{t+1} | H_t) = \Pr_\pi(A_{t+1} | S_t)$.  Let us denote this set by $\tilde{\policies}$.

\subsubsection{Baseline Agents}

We consider agents designed to learn policies within $\tilde{\policies}$.  While this class of policies is simple enough so that an agent could perform a nearly exhaustive search, we will restrict attention to approaches that can scale to settings involving much larger sets of aleatoric states.  Two simple agents of this kind will serve as baselines for comparison.  Let $\pi_\epsilon \in \tilde{\policies}$ be a policy that in the absence of a customer applies the slow mode, and otherwise samples slow or fast with probabilities $1-\epsilon$ and $\epsilon$.  Each of our baseline agents begins by executing $\pi_\epsilon$, with $\epsilon = 0$ -- that is, by applying the slow service mode over every timestep.  The first agent increases $\epsilon$ after gathering data over a long duration and using that data to estimate the arrival rate, if the estimate warrants increasing the service rate.  This agent's analysis is {\it static}, in the sense that it does not entail any experimentation and instead assumes the arrival rate will remain fixed.  The second agent additionally tries a small value of $\epsilon > 0$ for some duration in order to estimate the derivative $\mathrm{d} \lambda_{\pi_\epsilon} / \mathrm{d}\epsilon$ of average reward.  If this derivative is positive, it increases $\epsilon$.  One significant difference relative to the first agent is that, through its use of the derivative, this second agent anticipates the impact small increases in $\epsilon$ bear on the arrival rate.  As such, the second agent is representative of approaches used in the policy gradient literature, as discussed in \citep{sutton2018reinforcement} and references therein.

\subsubsection{Environment Dynamics}
\label{se:queue-environment}

We will study the behavior of agents given environment dynamics characterized by a specific observation probability function $\rho_*$, with the corresponding realized environment denoted by $e_* = (\actions, \observations, \rho_*)$.  We provide a detailed specification in Appendix \ref{se:service-station-appendix} and assume for the purposes of this analysis that $\Pr(\environment = e_*) > 0$.  In this environment, the customer arrival rate depends on the maximum service time experienced among the most recent dozen customers served.  The idea here is that long service times hurt reputation, which in turn reduces the number of customers seeking service.  Service times are impacted by the agent's choices: with the fast mode, service is always completed in a single timestep, while with the slow mode, service is completed over the next timestep with probability $1/2$.  Given our specification of $\rho_*$, the maximal average reward is $\$0.50$ per timestep.  This is achieved by applying the fast mode of service over every timestep, in which case each customer is served over a single timestep and a new customer arrives as soon as the previous one departs.

\subsubsection{Performance}
\label{se:queue-performance}

\begin{figure}[H]
\centering
\includegraphics[scale=0.2]{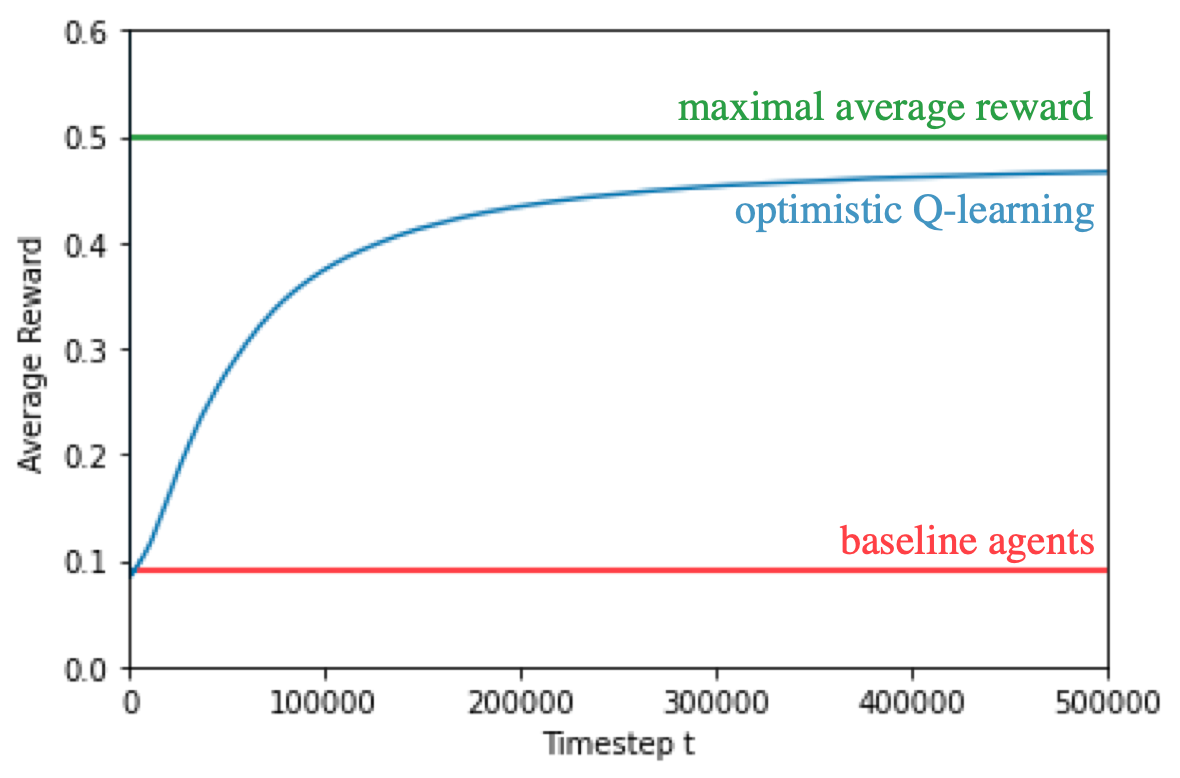}
\caption{Cumulative moving average rewards attained by an optimistic Q-learning agent, the maximal average reward, and the average reward attained by always applying the slow service mode, which approximates behavior of the baseline agents.}
\label{fig:agent-comparison}
\end{figure}

Figure \ref{fig:agent-comparison} plots cumulative moving average rewards attained by an optimistic Q-learning agent, which we will later present, averaged over two hundred independent simulations.  The figure also plots the maximum average reward and the average reward attained by always applying the slow service mode.  The baseline agents never choose to deviate from the slow service mode and therefore realize average reward close to the latter.

These results convey potential benefits of an agent designed to address general environments.  The optimistic Q-learning agent eventually figures out that its choices drive future arrival rates and based on this is able to improve its performance.  The baseline agents do not demonstrate that level of sophistication.

\section{Contributions and Related Literature}

This paper makes a range of contributions, innovating on formulation, framing of learning objectives, agent design, and performance analysis, as well as generating qualitative insights that can inform practical agent design.  In this section, we summarize these contributions and their relations to prior literature.

\subsection{Formulation}

Our formulation of agent-environment interactions is very general, involving a single stream of experience, without restrictive assumptions commonly made in the literature, as discussed in Section \ref{se:complex-environments}.  It is important for theoretical work to relax such assumptions if it is to inform the design of agents that can operate in complex real environments.  While our formulation bears close resemblance to those studied by \cite{mccallum1995instance,hutter2004universal,daswani2013q,daswani2014feature,lu2021reinforcement}, such formulations have not been a focus of work on provably efficient reinforcement learning.   Our work is the first to extend regret analysis tools to this setting.

\subsection{Framing of Learning Objectives}

In the literature on provably efficient reinforcement learning, it is common to study agent performance through regret analysis.  However, the manner in which regret bounds are typically framed does not suitably accommodate complex environments.  We develop concepts that allow us to frame meaningful learning objectives for such contexts.

\subsubsection{Averaging Time}

To intelligently choose between policies, an agent must assess their relative performance.  Regret bounds established in the literature typically reflect this requirement via dependence on statistics that bound the time required to assess an optimal policy.  Examples include, episode duration \citep{osband2013more,osband2019deep,azar2017minimax,jin2018q}, diameter \citep{jaksch2010near}, or span \citep{bartlett2012regal, ouyang2017learning, wei2020model}.  In a complex environment, the time required to assess an optimal policy can be intractably large or even infinite. As such, we will derive bounds that instead
depend on {\it reward averaging times} of policies in reference classes.

Let $\lambda_\pi(h,T)$ denote the expected average reward over $T$ timesteps starting at history $h$ so that $\lambda_\pi = \liminf_{T \rightarrow \infty} \lambda_\pi(H_0, T)$.  We define the reward averaging $\tau_\pi$ time of a policy $\pi \in \policies$ to be the smallest value $\tau \in [0,\infty)$ such that
\begin{equation}
\label{eq:averaging-time}
\left|\lambda_\pi(h,T) - \lambda_\pi\right| \leq \frac{\tau}{T},
\end{equation}
for all $h \in \histories$ and $T \geq 0$.  This is closely related to a concept introduced in \cite{kearns2002near}, which defines a notion of averaging time that is a function of a tolerance parameter associated with the error $|\lambda_\pi(h,T) - \lambda_\pi|$.  Our definition relies instead on a single scalar statistic $\tau_\pi$.  It is also worth noting that $\tau_{\pi_*}$, where $\pi_*$ is an optimal policy, is essentially equivalent to the notion of span introduced by \cite{bartlett2012regal}.

\subsubsection{Distortion}

A distinctive element of our formulation is in the agent's instantiation with an aleatoric state update function.  This serves to simplify the agent's experience by extracting useful features from history and enables productive behavior in arbitrarily complex environments.  In particular, instead of the number of environment states, our regret bounds will depend on the number of aleatoric states and the distortion incurred by using them to predict optimal discounted value.

Let $\phi(h)$ denote the aleatoric state that would be generated after experiencing a history $h \in \histories$.  For each discount factor $\gamma \in [0,1)$, history $h \in \histories$, and action $a \in \actions$, denote the optimal discounted action value by $Q_*^\gamma(h,a)$.  The discount factor $\gamma$ weights the reward realized after $k$ timesteps by $\gamma^k$ and can be thought of as prescribing an effective planning horizon of $\tau = 1/(1-\gamma)$.  We define the {\it distortion} for an effective planning horizon $\tau$ by
\begin{equation}
\label{eq:distortion-single-horizon}
    \Delta_\tau = \max_{(s,a) \in \states\times\actions} \left(\sup_{h \in \histories: \phi(h) = s} Q_*^\gamma(h,a) - \inf_{h \in \histories: \phi(h) = s} Q_*^\gamma(h,a)\right),
\end{equation}
where $\gamma = 1-1/\tau$.  This is the maximum difference between optimal action values across histories that lead to the same aleatoric state and offers a measure of error introduced when predicting optimal action values based on aleatoric state instead of history.  This sort of distortion measure has long been used in analysis of approximate dynamic programming algorithms that aggregate environment states \citep{whitt1978approximations,gordon1995stable,tsitsiklis1996feature,van2006performance}, though in this case we instead aggregate histories.

Our framing requires a stronger notion of distortion, defined by
\begin{equation}
\label{eq:distortion-sup-horizon}
\overline{\Delta}_\tau = \sup_{\tau' \geq \tau} \Delta_{\tau'}.
\end{equation}
This quantifies the accuracy with which aleatoric states can predict optimal action values for all planning horizons of duration $\tau$ or greater.  This distortion measure offers a useful statistic for characterizing performance of agents that are able to plan effectively over increasing horizons as data accumulates.

A limitation of our framing is in its use of a fixed aleatoric state update function.  While this is consistent with the manner in which some practical agents operate -- for example, the DQN agent of \citep{mnih2015human} takes its aleatoric state to be some number of recent video frames -- there is likely value to adapting the way in which aleatoric state is updated based on what is learned about the environment, which is encoded in the agent's epistemic state.  The MuZero agent \citep{schrittwieser2020muzero} does adapt its update function in this way.  That agent represents the update function in terms of a recurrent neural network, with weights adapted over time based on interactions with the environment.  Despite this limitation, our framing represents a significant step, advancing the mathematical literature in a direction that may inform future agent designs.

\subsubsection{Reference Classes}

We frame as agent design objectives a notion of competing effectively with policies from particular reference classes, with effectiveness measured through the lens of regret as a function of averaging times, distortions, and $\states$ and $\actions$.  As opposed to a single scalar objective, the spirit here is to offer a framework for studying trade-offs and to derive interpretable regret bounds that generate insight that can inform agent designers.  To understand this spirit, it may be helpful to draw an analogy with the field of optimization.  While optimization problems are framed in terms of precise scalar objectives, the design of optimization algorithms tends to be formulated in terms of measures of computational complexity and solution quality as a function of numbers of decision variables and constraints, as well as other salient problem characteristics.

One reference class we introduced earlier, denoted by $\tilde{\mathcal{P}}$, consists of all policies $\pi$ for which $\Pr_\pi(A_{t+1} | H_t) = \Pr_\pi(A_{t+1} | S_t)$.  In other words, these are the policies that select actions based on aleatoric state instead of history.  Ideally, the aleatoric state should suffice for predicting what the agent requires to make optimal decisions, in which case $\tilde{\mathcal{P}}$ would include an optimal policy.  Let $\tilde{\pi} \in \tilde{\mathcal{P}}$ be a policy for which $\lambda_{\tilde{\pi}} = \sup_{\pi \in \tilde{\mathcal{P}}} \lambda_\pi$.  We will think of the agent as trying to learn a high-performing policy from within $\tilde{\mathcal{P}}$, and as such, it is natural to expect that $\lambda_{\pi_{\rm agent}} \leq \lambda_{\tilde{\pi}}$.
As we will discuss in Section \ref{sec:adp}, there exist environments and aleatoric state dynamics such that $\lambda_* - \lambda_{\tilde{\pi}} \geq  \overline{\Delta}_{\tau_{\tilde{\pi}}}$, and consequently, if $\lambda_{\pi_{\rm agent}} \leq \lambda_{\tilde{\pi}}$, the average regret satisfies
\begin{equation}
\label{eq:ave-regret-lower-bound}
\liminf_{T\rightarrow \infty} \frac{\regret_{\mathcal{P}}(T)}{T} = \lambda_* - \lambda_{\pi_{\rm agent}} \geq \lambda_* - \lambda_{\tilde{\pi}} \geq \overline{\Delta}_{\tau_{\tilde{\pi}}}.
\end{equation}
In light of this fundamental limitation of the policy class $\tilde{\mathcal{P}}$, we frame as an objective optimizing the dependence of average regret on the distortion $\overline{\Delta}_{\tau_{\tilde{\pi}}}$.

The aforementioned objective calls for the agent to eventually compete effectively with the best policy among those that select actions based on aleatoric state.  A second objective we frame calls for the agent to attain that eventual level of performance quickly.  The time required depends on the time it takes to compare policies, which can be bounded by averaging times.  As discussed earlier, it is important to avoid dependence on the averaging time of an optimal policy as well as the number of environment states, each of which can be intractably large or even infinite in a complex environment.  We instead consider regret bounds that depend on the number of aleatoric states and the averaging time of $\tilde{\pi}$.  In particular, we consider regret bounds of the form
$$
\regret_{\mathcal{P}}(T) \leq {\tt foobar}(T, \states, \actions, \tau_{\tilde{\pi}}, \overline{\Delta}_{\tau_{\tilde{\pi}}}),
$$
where ${\tt foobar}$ is a metasyntactic function and, with some abuse of notation, we use $\states$ and $\actions$ to denote set cardinalities.  An understanding of how regret depends on the arguments can guide designs that more quickly learn to perform well relative to $\tilde{\pi}$.

We additionally consider, for each $\tau\geq 1$,  a reference class $\mathcal{P}_\tau = \{\pi \in \mathcal{P}: \tau_\pi \leq \tau\}$, consisting of policies with averaging times no greater than $\tau$.  For these classes, we consider bounds of the form
$$
\regret_{\mathcal{P}_\tau}(T) \leq {\tt foobar}(T, \states, \actions, \tau, \overline{\Delta}_{\tau}),
$$
for a different function ${\tt foobar}$.  Such bounds offer insight into how agents can quickly learn to perform well relative to policies with any particular averaging time.  An agent ought to be able to compete against policies in $\mathcal{P}_\tau$ within some time that grows with $\tau$, and such regret bounds reflect that relationship and draw attention to balancing associated trade-offs.

\subsection{Agent Design}

Our agent implements a variant of Q-learning \citep{watkins1989learning}.  Early analyses of Q-learning focused on asymptotic convergence guarantees under the assumption that the agent tries each action at each environment state infinitely often \citep{watkins1989learning,watkins1992q,tsitsiklis1994asynchronous,jaakkola1994convergence}.  More recently, research on Q-learning has merged with concepts from the literature on regret analysis, leading to provably efficient variations \citep{jin2018q,wei2020model}.  These {\it optimistic} Q-learning agents ensure a level of efficiency by using carefully chosen step sizes and perturbing action value updates to maintain optimistic estimates.  This merging presents an opportunity to bridge the efficient reinforcement learning literature with practical agent design, as Q-learning is more aligned with the state-of-the-art than other algorithms that have been studied in the mathematical literature.

While we build on this line of work to design a new optimistic Q-learning agent that is suitable for complex environments, our agent relies on several algorithmic innovations.  While the agents of \citep{jin2018q,wei2020model} maintain action values at each {\it environment state}, ours maintains action values at each {\it aleatoric state}.  Further, the algorithm of \cite{jin2018q} is designed for fixed-horizon episodic environments and that of \cite{wei2020model} operates with a fixed discount factor that depends on the horizon $T$.  Our algorithm is designed for general environments, and while it does make use of a discount factor, the discount factor increases over time to generate effective behavior over increasingly long planning horizons.  Further, while step sizes used in \citep{jin2018q,wei2020model} depend on the horizon $T$, our agent is designed to guide indefinitely rather than over a predetermined horizon $T$, and as such, uses step sizes that do not depend on $T$.

\subsection{Performance Analysis}

Critical contributions of this paper lie in our performance analysis.  While the results will be presented in Section \ref{sec:dynamic-horizon-algorithm}, here we discuss a few key implications.  Firstly, we establish that, if $\tau_{\tilde{\pi}}<\infty$, our agent attains average regret
\[
    \label{eq:regret-bound-simplified-2}
    \limsup_{T \rightarrow \infty} \frac{\regret_{\mathcal{P}}(T)}{T} = \lambda_* - \lambda_{\pi_{\rm agent}} \leq 4 \overline{\Delta}_{\tau_{\tilde{\pi}}}.
\]
This is exactly four times the lower bound of \eqref{eq:ave-regret-lower-bound}.  It is also interesting to relate this upper bound to Theorem \ref{theorem:adp-fail} in Section \ref{sec:adp}, which indicates that, for all $\epsilon > 0$, there exists an environment, a set of aleatoric states, an aleatoric state update function and a reward function, such that particular approximate dynamic programming (ADP) methods one might apply (e.g., \cite{whitt1978approximations,gordon1995stable,tsitsiklis1996feature,munos2008finite}) yield a policy $\pi_{\rm ADP}$ for which $\lambda_* - \lambda_{\pi_{\rm ADP}} \geq \tau_{\tilde{\pi}} \overline{\Delta}_{\tau_{\tilde{\pi}}} - \epsilon$, which is generally far worse that $4 \overline{\Delta}_{\tau_{\tilde{\pi}}}$.  Further, \cite{van2006performance} suggests that a {\it temporal-difference fixed point} would yield a policy $\pi_{\rm TD}$ that satisfies $\lambda_* - \lambda_{\pi_{\rm TD}} \leq \overline{\Delta}_{\tau_{\tilde{\pi}}}$, but it is not known whether such a fixed point can be determined by a computationally tractable algorithm. 
It is intriguing that our agent -- which is computationally tractable and itself based on a temporal-difference method -- attains average regret within a factor of four of that.

Specialized to the case where the distortion $\overline{\Delta}_{\tau_{\tilde{\pi}}} = 0$, our analysis implies the following:
\[
    \label{eq:regret-bound-simplified}
    \regret_{\mathcal{P}}(T) \lesssim \left( \sqrt{\states\actions} + \tau_{\tilde{\pi}}\right) T^{4/5} + \states\actions T^{1/5} + \tau_{\tilde{\pi}}^5,
\]
where $\lesssim$ indicates omission of constant and poly-logarithmic factors.  In this case, since aleatoric states enable exact predictions of optimal value, the regret grows sublinearly in $T$, meaning that the agent eventually learns a globally optimal policy.  The dependence on $T$ is worse than the usual $T^{1/2}$ scaling, which appears in results pertaining to episodic environments (\cite{jin2018q, zhang2020reinforcement}).  In our formulation, a $T^{1/2}$ scaling is unachievable without additional problem-dependent terms in the regret bound that scale exponentially with $\states$ and $\actions$ (\cite{jaksch2010near, wei2020model}).  It is worth noting that, while \cite{wei2020model} considers an average reward objective, though with zero distortion, and provides a regret bound that scales with $T^{2/3}$ rather than $T^{4/5}$, the algorithm crucially relies on knowledge of a fixed duration $T$.  Our agent and analysis can also be modified to attain a $T^{2/3}$ scaling given a fixed duration $T$.

Combining \eqref{eq:regret-bound-simplified-2} and \eqref{eq:regret-bound-simplified}, we can see that besides $\states, \actions$ and $T$, the bound only depends on $\tau_{\tilde{\pi}}$, the reward averaging time of the best policy in the reference class.  Previous regret bounds for tabular reinforcement learning scale with the number of states or the reward averaging time of an optimal policy.  In a complex environment, these quantities can be arbitrarily large or infinite.  Interestingly, our bound ensures that the agent is able to learn efficiently in spite of that. 

We further establish that, for all $\tau\geq 1$, 
\[
    \label{eq:regret-bound-simplified-3}
    \regret_{\mathcal{P}_\tau}(T) \lesssim \left( \sqrt{\states\actions} + \tau\right) T^{4/5} + \states\actions T^{1/5} + \tau^5 + \overline{\Delta}_\tau T.
\]
Recall that $\mathcal{P}_\tau$ is the class of policies with reward averaging times no greater than $\tau$ and $\regret_{\mathcal{P}_\tau}(T)$ quantifies regret relative to that class.  This bound offers insight into how, over time, the agent can learn to perform competitively against policies with larger reward averaging times.  To understand this, let us focus on a special case where $\Delta_\tau = 0$ for all $\tau$.  In this case, the bound implies that, for all $\epsilon \in (0,1)$, setting $\tau = \epsilon  T^{1/5}$, 
\[
\limsup_{T \rightarrow \infty} \frac{\regret_{\mathcal{P}_\tau}(T)}{T} \lesssim \epsilon.
\]
Hence, for sufficiently large $T$, the agent's average reward approximates that of the best policy with reward averaging time no greater than $\epsilon T^{1/5}$.

\subsection{Qualitative Insights}

While it shares elements common to state-of-the-art agents, our agent is far simpler.  Our motivation was not to produce another state-of-the-art agent, but rather to offer a context amenable to analyses that can inform design of future state-of-the-art agents.  We now discuss some key insights supported by our results.  

First of all, our results demonstrate that it is possible for an agent to operate effectively within a tractable time frame through a single endless stream of interactions with an arbitrarily complex environment. 
Previous results either rely on the fact that the environment mixes in a modest amount of time \citep{jin2018q, jaksch2010near, zhang2020reinforcement} or that the horizon $T$ of operation is fixed and known to the agent \citep{wei2020model}.  Further, previous results focus on MDPs, and while there has also been related work on POMDPs \citep{jafarnia2021online, kara2020near, subramanian2020approximate}, those results are relevant only when there is a tractable number of environment states.  Our bounds do not depend on the environment's mixing time or number of states.  Among other things, our results imply that an agent can perform well even in an environment that is so complex that the performance of an optimal policy would take forever to estimate.

Secondly, we are the first to establish that an algorithm with average regret bounded by a constant multiple of distortion approaches such asymptotic performance within a tractable time frame.  An example in \citep{van2006performance} implies that certain common ADP algorithms, which require that environment dynamics be known, do not output a policy $\pi\in\tilde{\policies}$ such that $\lambda_* - \lambda_\pi$ is within a constant multiple of $\overline{\Delta}$.  Indeed, previous analyses of ADP algorithms instead bound $\lambda_* - \lambda_\pi$ by a multiple of $\tau\overline{\Delta}$, where $\tau$ is some notion of averaging time that depends on environment complexity \citep{whitt1978approximations, gordon1995stable,tsitsiklis1996feature}.  This scaling by $\tau$ is far worse than a constant, with $\tau$ becoming arbitrarily large in complex environments.  In real environments, it is impractical to attain zero distortion, and therefore, some degree of impact on performance is inevitable.  Our result offers insight into how to avoid scaling by $\tau$.

An intriguing aspect of our agent design is that the effective planning horizon increases with time, allowing the agent to eventually optimize performance over arbitrarily long horizons.  Our agent's effective planning horizon scales with $t^{1/5}$, and this rate leads to our regret bound.  The notion that planning may benefit from restricting the effective horizon based on the quantity of data gathered has also been observed by \cite{jiang2015dependence}.

Our regret bounds depend on the distortion induced by a fixed aleatoric state update function.  However, some state-of-the-art agents leverage the ability of neural networks to adapt this update function \citep{nachum2018near, schrittwieser2020muzero}.  While our results do not directly address such adaptation, they do offer insight into the way in which that can influence agent performance.

\section{Value Functions}
\label{sec:value-functions}

Central to the theory of MDPs are value functions.  While value functions are typically considered to be functions of environment state, we consider instead functions of history.  In this section we define these value functions and characterize them as solutions to Bellman equations.

Throughout this section, we consider a fixed discount factor $\gamma \in [0,1)$ and environment $\environment=(\actions, \observations, \rho)$.  
To simplify notation, we will use $(h, a, o)$ to denote the history generated by concatenating action $a$ and observation $o$ to history $h$.  For each $a\in\actions$, we define an $\histories\times\histories$ transition matrix $P_a$, with entries
\[
    P_{ahh'} = 
    \begin{cases}
        \rho(o|h,a) & \text{if $h'=(h,a,o)$}\\
        0 & \text{otherwise}
    \end{cases},
\]
for each $h,h' \in \histories$.  Similarly, for each policy $\pi\in\policies$, we define a transition matrix $P_\pi$, with
\[
    P_{\pi hh'} = \sum_{a\in\actions} \Big(\pi(a|h)\cdot P_{ahh'}\Big), \quad
    \forall h, h'\in\histories,
\]
for each $h,h' \in \histories$.  Further, for each action $a \in \actions$ and policy $\pi \in \policies$, let $\overline{r}_{a}$ and $\overline{r}_\pi$ be $\histories$-dimensional vectors, with components given by
\[
    \overline{r}_{ah} = \sum_{o\in\observations} \Big(\rho(o|h,a)\cdot r\big(\phi(h), a, o\big)\Big) \qquad \text{and} \qquad  
    \overline{r}_{\pi h} = \sum_{a\in\actions} \Big(\pi(a|h)\cdot \overline{r}_{ah}\Big).
\]

For each policy $\pi \in \policies$, let
\[
    V_\pi^\gamma(h) = \sum_{t=0}^\infty \Big(\gamma^t\cdot \big(P_\pi^t \overline{r}_\pi\big) (h)\Big)
    \qquad \text{and} \qquad
    Q_\pi^\gamma(h, a) = 
    \overline{r}_{ah} + \sum_{h'\in\histories} \Big(P_{ahh'}\cdot V_\pi^\gamma(h') \Big).
\]
These functions represent expected discounted rewards starting at history $h$ if either all subsequent actions are selected by $\pi$ or only after an action $a$ is executed.
By taking the supremum over policies, we obtain optimal values:
\[
    V_{*}^\gamma(h) = \sup_{\pi\in\policies} V_\pi^\gamma(h) \qquad \text{and} \qquad  Q_{*}^\gamma(h, a) =  \sup_{\pi\in\policies} Q_\pi^\gamma(h, a),\quad \forall h\in\histories, a\in\actions.
\]
The following proposition, which follows from Proposition 2.1.1 in \cite{bertsekas2018abstract}, characterizes $V_*^\gamma$ and $Q_*^\gamma$ as unique solutions among the set of bounded functions to the Bellman equations.
\begin{proposition}
\label{lemma:bellman}
The pair $(V_*^\gamma, Q_*^\gamma)$ uniquely solves the system of equations
$$\begin{array}{ll}
V(h) = \max_{a'\in\actions} Q(h, a') \qquad & \forall h\in\histories \\
Q(h, a) = \rbar_{h,a} + \gamma\cdot\sum_{h'\in\histories} \Big(P_{ahh'}\cdot V(h') \Big) \qquad & \forall h\in\histories, a\in\actions.
\end{array}$$
among all pairs of bounded functions $V:\histories\rightarrow\mathbb{R}$ and $Q:\histories\times\actions\rightarrow \mathbb{R}$.
\end{proposition}

We close this section with an important lemma, which ties together three concepts relating to a policy $\pi$: the long-term expected average reward $\lambda_\pi$, the reward averaging time $\tau_\pi$, and the discounted value function $V_\pi^\gamma$.  The lemma closely resembles Theorem 4.1 in \cite{defarias2006cost}, and we omit the proof.
\begin{restatable}{lemma}{mixingtimelemma}
\label{lemma:mixing-time}
For all $\pi\in\policies$, $h\in\histories$ and $\gamma\in[0,1)$,
$\left|V_\pi^\gamma(h) - \frac{\lambda_\pi}{1-\gamma} \right|\leq \tau_\pi$.
\end{restatable}
This result establishes that the reward averaging time bounds the difference between discounted value $V_\pi^\gamma(h)$ and average reward $\lambda_\pi$ scaled by the effective horizon $1/ (1-\gamma)$.

\section{Agent Design and Performance Analysis}
\label{sec:dynamic-horizon-algorithm}

In this section we present and study our optimistic Q-learning agent.  Similarly with agents that have demonstrated success in large-scale simulations, ours learns to predict action values.  However, rather than a neural network representation, our agent maintains a lookup table containing one prediction per (aleatoric) state-action pair.  Actions are selected greedily with respect to these predictions.  Upon each observation, the agent incrementally adjusts the prediction assigned to its previous state-action pair based on a temporal difference.

The agent predicts discounted value.  However, in order to eventually maximize average reward, the associated discount factor increases over time and approaches one.  The idea is for the agent to plan, at any given time, over a particular effective horizon.  This horizon increases as the agent gathers more data, which enables planning over longer horizons with greater confidence.

\subsection{Discounted $Q$-Learning}
\label{subsubsec:discounted-q-learning}

As a prelude to our primary agent, we introduce a simpler one that serves didactic purposes.  This simper agent plans over a fixed effective horizon $\tau$ and is designed to operate over a fixed duration $T \gg \tau$, with both these variables required as input when instantiating the agent.  In particular, the agent executes Algorithm \ref{alg:q-learning-update} ({\tt discounted\_q\_learning}). The effective horizon $\tau$ prescribes a discount factor $\gamma = 1-1/\tau$.  The agent starts with an initial aleatoric state $S_0$.  Over each timestep, the agent increments the visitation count $N(s,a)$, computes the next aleatoric state $s' = \falea(s, a, o)$, and updates the prediction $Q(s, a)$ via a discounted $Q$-learning iteration, with discount factor $\gamma = 1-1/\tau$.

\begin{algorithm}
    \begin{tabular}{lll}
    \textbf{Input:} & $S_0$ & initial aleatoric state\\
        & $\falea$ & aleatoric state update function\\
        & $r$ & reward function\\
        & $\tau$ & effective planning horizon \\
        & $T$ & duration of operation \\
    \end{tabular}
    \begin{algorithmic}[1]
    \STATE $\gamma\leftarrow 1 - 1/\tau$
    \STATE $\beta \leftarrow \tau^{3/2} \cdot 4\sqrt{\log (2T^2)}$
    \STATE $s\leftarrow S_0$
    \STATE $Q(\cdot, \cdot)\leftarrow \tau$
    \FOR{$t=1,2,\dots, T$}
        \STATE $a\leftarrow \mathtt{sample\_unif}\big(\argmax_{a'\in\actions}Q(s,a')\big) $
        \STATE execute action $a$ and register observation $o$ 
        \STATE $N(s,a) \leftarrow N(s,a) + 1$
        \STATE $\alpha \leftarrow \frac{1 + 2\tau}{N(s,a) + 2\tau}$
        \STATE $s' \leftarrow \falea(s,a,o)$
        \STATE $Q(s,a)  \leftarrow Q(s,a) + \alpha \cdot \left(r(s,a,o) + \gamma \cdot \max_{a' \in \actions} Q(s', a') - Q(s,a) + \frac{\beta}{\sqrt{N(s,a)}} \right)$
        \STATE $Q(s,a) \leftarrow \min(Q(s,a), \tau)$
        \STATE $s \leftarrow s'$
    \ENDFOR
    \end{algorithmic}
    \caption{\tt discounted\_q\_learning}
    \label{alg:q-learning-update}
\end{algorithm}\par

The Q-learning update of Line 11 adjusts the action value in response to a temporal difference.  Two elements of this update warrant further discussion.  One is the step size $\alpha$, which is given by $(1 + 2\tau) / (N(s,a) + 2\tau)$.  This step size sequence is adapted from that used in \cite{jin2018q} and has a number of desirable properties, as will be established in Lemma \ref{lemma:learning-rate}.  In particular, these properties ensure that estimation errors do not accumulate exponentially as the agent updates action values.  A second key element is the optimistic boost added to the temporal difference, which is given by $\beta/\sqrt{N(s,a)}$.  This term injects optimism to ensure that predictions are likely to be optimistic, in terms of dominating $Q_*^\gamma$.  As the number of visits $N(s,a)$ to a state-action pair increases, uncertainty around its prediction decreases, and this is reflected in the denominator $\sqrt{N(s,a)}$.


Let $\pi_{\rm agent}^{\tau,T} \in \policies$ be the policy implemented by an agent that executes Algorithm \ref{alg:q-learning-update} with effective planning horizon $\tau$ and operation duration $T$.   Let the regret relative to a reference policy $\pi \in \policies$ experienced by this agent over $T$ timesteps be denoted by
\[
    \regret^{\tau}_\pi(T) = \E_{\pi_{\rm agent}^{\tau,T}}\left[ \sum_{t=0}^{T-1} \big( \lambda_{\pi} - R_{t+1} \big) \Big| \environment\right].
\]
Recall that $\tau_\pi = \inf_{T \geq 0} \left|\lambda_\pi(h,T) - \lambda_\pi\right|$ is the reward averaging time of policy $\pi$ and 
$$\Delta_\tau = \max_{(s,a) \in \states\times\actions} \left(\sup_{h \in \histories: \phi(h)=s} Q_*^\gamma(h,a) - \inf_{h \in \histories: \phi(h)=s} Q_*^\gamma(h,a)\right),$$
where $\gamma = 1-1/\tau$, is the distortion introduced in predicting the optimal value over effective horizon $\tau$ based on the aleatoric state instead of history.
We have the following regret bound.
\begin{restatable}{theorem}{discountedtheorem}
\label{theorem:average-reward-discounted-algorithm}
For all $\tau\geq 1, T \geq 1$ and $\pi\in\policies$, we have
\begin{eqnarray}
    \regret_\pi^{\tau}(T)
    \leq 24\tau^{3/2} \cdot\sqrt{\states\actions T\log (2T^2)} + \Big[3\Delta_\tau + \tau_{\pi}/\tau\Big]\cdot T + \big[\states\actions + 5+2\log(T)\big]\cdot \tau. \nn
\end{eqnarray}
\end{restatable}

While Algorithm \ref{alg:q-learning-update} requires the effective planning horizon $\tau$ and the duration $T$ of operation as input, we establish in Section \ref{subsubsec:expanding-the-horizon} a regret bound for a more sophisticated agent that does not require $\tau$ or $T$ as input.  The agent relaxes the need for these parameters by operating with an effective planning horizon that increases over time.

\subsection{Growing the Horizon}
\label{subsubsec:expanding-the-horizon}

Rather than targeting fixing the duration of operation and the effective planning horizon, as done by Algorithm \ref{alg:q-learning-update} ({\tt discounted\_q\_learning}), we can design an agent that operates effectively over any duration by planning over a growing horizon.  We now study our primary agent, which executes Algorithm \ref{alg:dynamic-horizon-q-learning} ({\tt growing\_horizon\_q\_learning})  to accomplish this.  The agent is instantiated with only three inputs: an initial aleatoric state, an aleatoric state update function, and a reward function.  It is worth noting that the agent interacts with the environment through a single stream of experience, with no resets or reinitialization of the aleatoric state.  While the Q-learning update of Line 14 is looks identical to that of Algorithm \ref{alg:q-learning-update}, the effective horizon $\tau$ -- and thus, the discount factor $\gamma$ -- and optimism coefficient $\beta$ now change over time.

\begin{algorithm}
    \begin{tabular}{lll}
    \textbf{Input:} & $S_0$ & initial aleatoric state\\
        & $\falea$ & aleatoric state update function\\
        & $r$ & reward function\\
    \end{tabular}
    \begin{algorithmic}[1]
    \STATE $s\leftarrow S_0$
    \STATE $Q(\cdot, \cdot) \leftarrow 1,\quad  N(\cdot, \cdot) \leftarrow 0$
    \FOR {$t=1,2,\dots$}
        \STATE $\tau\leftarrow {\tt foo}_1(t)$
        \STATE $\beta\leftarrow {\tt foo}_2(t)$
        \STATE $Q(\cdot, \cdot) \leftarrow Q(\cdot, \cdot) +  {\tt foo}_3(t)$
        \STATE $N(\cdot, \cdot) \leftarrow N(\cdot, \cdot) \cdot  {\tt foo}_4(t)$
        \STATE $a\leftarrow \mathtt{sample\_unif}\big(\argmax_{a'\in\actions}Q(s,a')\big) $
        \STATE execute action $a$ and register observation $o$
        \STATE $N(s,a)\leftarrow N(s,a) + 1$
        \STATE $\alpha \leftarrow \frac{1 + 2\tau}{N(s,a) + 2\tau}$
        \STATE $s' \leftarrow \falea(s,a,o)$
        \STATE $\gamma \leftarrow 1 - 1/\tau$
        \STATE $Q(s,a)  \leftarrow Q(s,a) + \alpha \cdot \left(r(s,a,o) + \gamma \cdot \max_{a' \in \actions} Q(s', a') - Q(s,a) + \frac{\beta}{\sqrt{N(s,a)}} \right)$
        \STATE $Q(s,a) \leftarrow \min(Q(s,a), \tau)$
        \STATE $s \leftarrow s'$
    \ENDFOR
    \end{algorithmic}
    \caption{\tt growing\_horizon\_q\_learning}
    \label{alg:dynamic-horizon-q-learning}
\end{algorithm}\par

The algorithm calls subroutines ${\tt foo}_1$ through ${\tt foo}_4$, which govern evolution of the effective planning horizon $\tau$ the optimism coefficient $\beta$, and suitably adjust action values $Q$ and visitation counts $N$ in tandem with changes in $\tau$ and $\beta$.  In particular,
\begin{itemize}
    \item ${\tt foo}_1(t)$ prescribes the effective planning horizon;
    \item ${\tt foo}_2(t)$ prescribes the optimism coefficient;
    \item ${\tt foo}_3(t)$ increases all action values so that they remain optimistic as the effective planning horizon increases;
    \item ${\tt foo}_4(t)$ deemphasizes less recent temporal differences, which were based on a substantially different discount factor.
\end{itemize}

A sequence of {\it change points}, beginning with $T_0=1$ and continuing with $T_k = 20\cdot 2^{k-1}$ for $k =1,2,3,\ldots$, underlie these functions.  To specify the functions, it is helpful to define notation for the {\it most recent change point} at each time $t$.  In particular, with the index of the most recent change point given by $k_t = \max\{k \geq 0: T_k \leq t\}$ if $t>0$ and $k_0=0$, the most recent change point is $T_{k_t}$.  The first of these functions, which provides the effective planning horizon $\tau$, is
\[
    {\tt foo}_1(t) = T_{k_t}^{1/5}. \label{eq:foo-1}
\]
The optimism coefficient $\beta$ is similarly updated at changed points according to
\[
    {\tt foo}_2(t) = 4T_{k_t}^{3/10} \sqrt{\log (2T_{k_t}^2)}.
\]  
Note that $T_{k_t}^{3/10} = {\tt foo}^{3/2}_1(t)$, so the optimism coefficient scales with the effective planning horizon raised to a power of $3/2$ times a logarithmic term.  To ensure that the action values remain, they are incremented by the same amount as the effective horizon; this is accomplished by
\[
    {\tt foo}_3(t) =  T_{k_t}^{1/5}-T_{k_{t-1}}^{1/5}.
    \label{eq:foo-3}
\]
Finally, to simplify analysis, we reset state-action counts at change points by multiplying them by
\[
    {\tt foo}_4(t) ={\bf 1}(T_{k_t} = T_{k_{t-1}}). \label{eq:foo-4}
\]
The count becomes one upon the next visit to any state-action pair, and the resulting step size $\alpha = 1$ replaces the action value with the temporal difference, effectively forcing the agent to forget all experience preceding the change point.  It is important to note that these choices of ${\tt foo}_1$ through ${\tt foo}_4$ were designed to facilitate analysis rather to produce the most effective agent.  We will discuss alternative choices in the next section that may improve performance.

We denote by $\pi_{\rm agent}$ the policy executed by Algorithm \ref{alg:dynamic-horizon-q-learning} with the subroutines specified above.  Recall that $\regret_\pi(T)$ is the regret experienced by $\pi_{\rm agent}$ relative to a reference policy $\pi \in \policies$ and that
$\overline{\Delta}_{\tau_\pi} = \sup_{\tau\geq \tau_\pi} \Delta_\tau$ is the maximum distortion over effective horizons equal to or exceeding the reward averaging time $\tau_\pi$.
The following theorem is the main theoretical result of this paper.
\begin{restatable}{theorem}{averagerewardtheorem}
\label{theorem:main}
For all $\pi\in\policies$ and $T \geq 1$, 
\begin{equation}
    \regret_{\pi}(T)
    \leq \left(120 \sqrt{\states \actions \log (2T^2)} + 5 \tau_{\pi} \right) T^{4/5} + 3 \overline{\Delta}_{\tau_\pi} T + \left(54 \states \actions + 18\log(T)\right) T^{1/5}+ 2 \tau_{\pi}^5.   \nn
\end{equation}
\end{restatable}

Algorithm \ref{alg:dynamic-horizon-q-learning} can be viewed as operating over a sequence of episodes, delineated by change points and with the effective planning horizon and optimism coefficient fixed over each.  As such, it can be thought of as instantiating and applying \ref{alg:q-learning-update} over each episode.  Despite that, Theorem \ref{theorem:main} is not follow directly from Theorem \ref{theorem:average-reward-discounted-algorithm}.  The reason is that the latter applies to an agent that begins with an empty history, whereas an agent that is instantiated at some change point does not.  Theorem \ref{theorem:average-reward-discounted-algorithm} in Appendix \ref{subsec:discounted-to-average} bridges this gap, offering a generalization to Theorem \ref{theorem:average-reward-discounted-algorithm} that applies to an agent starting with an arbitrary history $h\in\histories$ so long as its aleatoric state is initialized to $\phi(h)$.

Two corollaries of Theorem \ref{theorem:main} facilitate interpretation of its implications.  The first characterizes regret relative to reference classes $\policies_\tau$, each of which consists of policies for which reward averaging times do not exceed $\tau$.  Since $\regret_{\policies_\tau}(T) = \sup_{\pi \in \policies_\tau} \regret_\pi(T)$, we have the following corollary.
\begin{corollary}
\label{cor:avg-time-upper-bound}
For all $\tau\geq 1$ and $T\geq 1$, 
\begin{equation}
    \regret_{\mathcal{P}_\tau}(T)
    \leq \left(120 \sqrt{\states \actions \log (2T^2)} + 5 \tau \right) T^{4/5} + 3 \overline{\Delta}_{\tau} T + \left(54 \states \actions + 18\log(T)\right) T^{1/5}+ 2 \tau^5.   
\end{equation}
\end{corollary}
It follows from this corollary that, for all $\epsilon>0,\tau \geq 1$ and some polynomial ${\tt poly}(\cdot, \cdot, \cdot)$, the agent attains average reward within $3\overline{\Delta}_\tau + \epsilon$ of $\sup_{\pi \in \policies_\tau} \lambda_\pi$ within $\tau^5 \cdot {\tt poly}(\states,\actions,1/\epsilon)$ timesteps.  Hence, within time that scales with $\tau^5$, the agent attains average reward competitive with any policy with reward averaging time $\tau$.  Also implicit in this observation is that, over time, the agent becomes competitive with policies that require longer times to evaluate.

A second corollary bounds regret relative to the optimal average reward $\lambda_*$.  This follows from Theorem \ref{theorem:main} and our next lemma, which is a consequence of Corollary 5.1 of \citep{van2006performance}.
\begin{restatable}{lemma}{aleatoricvsglobal}
    \label{lemma:aleatoric-vs-global}
    For all $\tau\geq 1$, $\lambda_{*} - \lambda_{\tilde{\pi}} \leq \overline{\Delta}_\tau$.
\end{restatable}
Applying Lemma \ref{lemma:aleatoric-vs-global} and taking $\pi$ to be $\tilde{\pi}$, we arrive at the following corollary to Theorem \ref{theorem:main}.
\begin{corollary}
\label{cor:global-optimum}
For all $T\geq 1$, 
\begin{equation}
    \label{eq:cor:global-optimum}
    \regret_{\mathcal{P}}(T)
    \leq \left(120 \sqrt{\states \actions \log (2T^2)} + 5 \tau_{\tilde{\pi}} \right) T^{4/5} + 4 \overline{\Delta}_{\tau_{\tilde{\pi}}} T + \left(54 \states \actions + 18\log(T)\right) T^{1/5}+ 2 \tau_{\tilde{\pi}}^5.   
\end{equation}
\end{corollary}
This corollary conveys another intriguing property of our result: the agent approaches its asymptotic performance in time that scales with $\tau_{\tilde{\pi}}^5$.  In particular, this time does not depend on the reward averaging time of an optimal policy, which in a complex environment could be intractably large or even infinite.  The dependence is instead on the reward averaging time of $\tilde{\pi}$, which is determined by aleatoric, rather than environment, state dynamics.

\subsection{Scheduling Schemes}
\label{subsec:scheduling-schemes}

The functions ${\tt foo}_1$ through ${\tt foo}_4$ prescribe schedules for adjusting the effective planning horizon, the optimism coefficient, and value and count functions.  The particular choices specified in the previous section as part of Algorithm \ref{alg:dynamic-horizon-q-learning} were designed to facilitate regret analysis.  Indeed, their irregular structure, with abrupt adjustments occurring at particular change points, was introduced solely to simplify analysis by partitioning the stream into episodes.  More natural choices involving ``smooth'' schedules may substantially improve realized performance while satisfying similar or improved regret bounds.  Further, the rate at which the effective horizon grows with time plays an important role, and a rate of $t^{1/5}$ may be onerously slow, requiring a very long time to develop plans that span reasonable horizons.

\begin{figure}[H]
\centering
\includegraphics[scale=0.2]{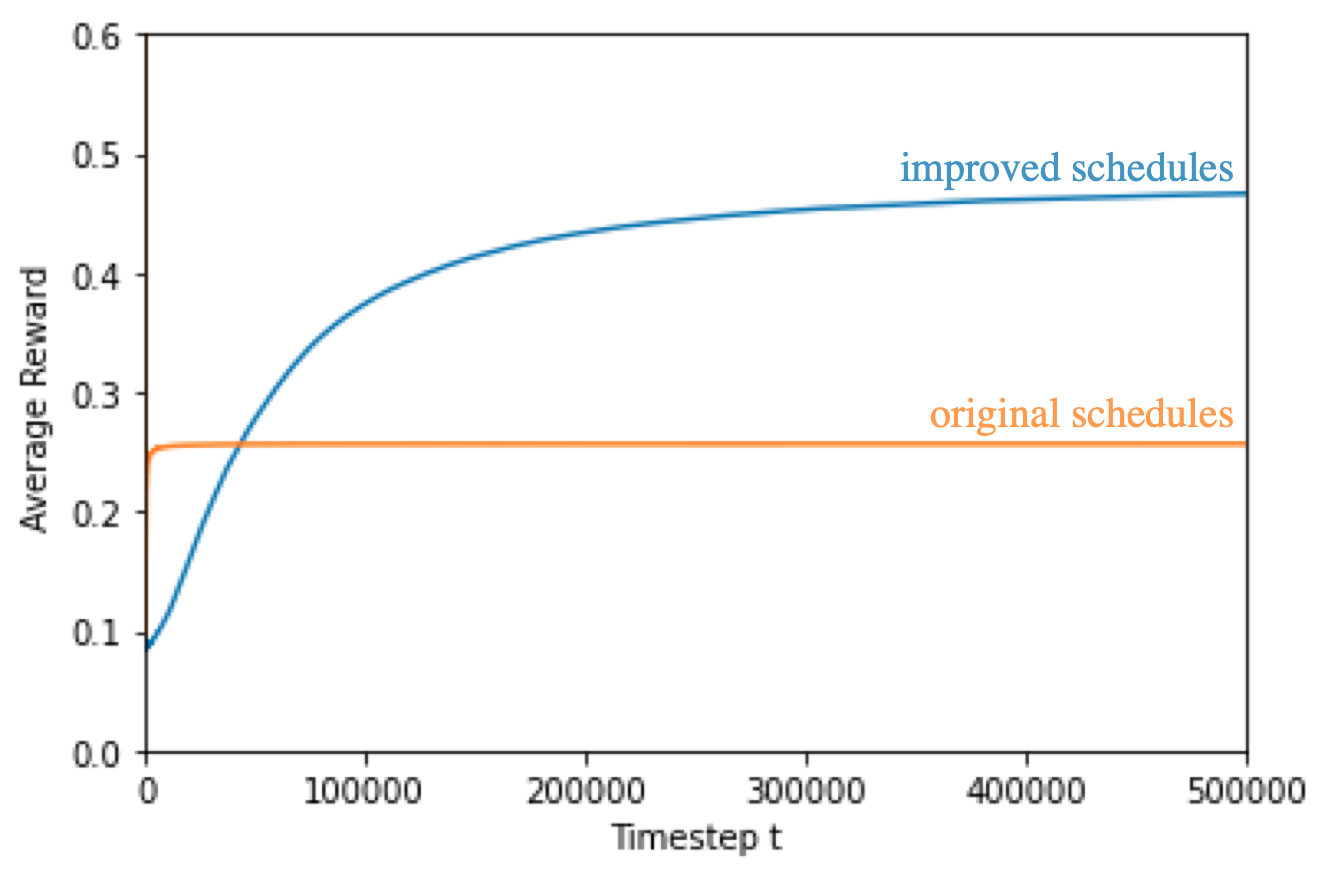}
\caption{Performance of Algorithm \ref{alg:dynamic-horizon-q-learning} with its original schedules versus improved smooth schedules.}
\label{fig:better-schedules}
\end{figure}

To illustrate the importance of these schedules, let us revisit the service rate control example of Section \ref{se:service-station}.  Simulation results reported in that section, which demonstrated the capability of optimistic Q-learning to improve performance over time, made use of particular smooth schedules:
\begin{align*}
{\tt foo}_1(t) =& 1.5 t^{1/5}, \\
{\tt foo}_2(t) =& 0.44 t^{3/10} \sqrt{\log(2 t^2)}, \\
{\tt foo}_3(t) =& 1.5 (t^{1/5} - (t-1)^{1/5}), \\
{\tt foo}_4(t) =& 1.
\end{align*}
Note that, while it may be beneficial to modify the rate at which the planning horizon grows, for the purposes of our current study, we retain the $t^{1/5}$ rate and only tune other aspects of the schedules.  Figure \ref{fig:better-schedules} compares results reported in Section \ref{se:queue-performance} against the schedules of Algorithm \ref{alg:dynamic-horizon-q-learning}.  Each plot represents an average over two hundred simulated trajectories.  While the latter agent eventually improves performance, that requires a very long time due to its impractical schedules.

It may be surprising that Algorithm \ref{alg:dynamic-horizon-q-learning} performs so poorly despite satisfying regret bounds of the previous section.  Indeed, as is common to mathematical results on efficient reinforcement learning, such regret bounds tend to be very weak.  They typically do not offer accurate predictions of realized performance, and given the level of inaccuracy, they do not offer precise guidance on agent design.  However, these bounds and the analyses that lead to them, can be useful for developing qualitative understanding and insights, as we have discussed in earlier sections.

\section{Closing Remarks}

We presented and studied a simple agent that through a general agent-environment interface interacts over a single stream of experience.  Our results bound regret realized by the agent.  These bounds bear implication on asymptotic performance and the rate at which the agent approaches that level of performance.  Importantly, these bounds do not depend on the number of environment states or their mixing time.  One interesting insight that emerges involves the relation between the agent's effective planning horizon and its duration $T$ of past experience: the agent plans effectively over a horizon that grows with $T^{1/5}$.  There are a number of directions in which the results in this work can be strengthened or extended.  We will discuss a few in this section.

Our agent uses a particularly simple representation for the action value function, comprised of a fixed, prespecified aleatoric state update function and a lookup table over aleatoric states.  State-of-the-art agents adapt the aleatoric state update function based on the agent's experience and generalize over aleatoric states, typically by using a neural network instead of a lookup table, and these extensions allow for much larger aleatoric state spaces and can greatly improve performance.  

The agent that we analyze discards all previous experience whenever the effective planning horizon is increased.  This is impractical and done only to facilitate analysis.  It ought to be possible to analyze a variation that more gradually phases out the influence of past data, along the lines discussed in Section \ref{subsec:scheduling-schemes}.

To maximize long-term average reward, it may be natural for the agent to learn the differential value functions directly, as is studied in \cite{wan2020learning}, rather than discounted value functions, as does our agent.  However, an open issue is whether an agent can explore the environment efficiently when doing so.  We believe that this is a problem that is worth further investigation.

Finally, we suspect that the $T^{4/5}$ term in our regret bound, which reflects the rate at which the agent approaches its asymptotic performance, is not fundamental and can be improved with a better agent design and a more nuanced analysis.  We note that this dependence stems from the subroutines $\mathtt{foo}_1$ \eqref{eq:foo-1} through $\mathtt{foo}_4$ \eqref{eq:foo-4} that the agent uses to adjust the planning horizon.  As we mentioned in Section \ref{subsubsec:expanding-the-horizon}, these settings induce an effectively planning horizon that grows with $T^{1/5}$, suggesting that it takes time $\tau^5$ to plan effectively over a horizon of length $\tau$.  We conjecture that there exists an environment in which any agent requires time $\tau^3$ to do this, which would translate to a $T^{2/3}$ instead of $T^{4/5}$ term in the regret lower bound.  Such a lower bound could shed light on the limits of learning and could offer useful insight to agent designers on how long the agent ought to plan, given the duration of past experience.

\section*{Acknowledgements}

We thank Satinder Singh and John Tsitsiklis for stimulating discussions and helpful feedback.  We also thank Alex Cloud for pointing out a mistake in a previous version of this paper.  Financial support from Army Research Office (ARO) grant W911NF2010055 is gratefully acknowledged.  Shi Dong was also supported by the Herb and Jane Dwight Stanford Graduate Fellowship.

\appendix

\section{Probabilistic Framework}
\label{se:probability}

In this appendix, we define our probabilistic framework and notation. We will define all random quantities with respect to a probability space $(\Omega, \F, \Pr)$.  The probability of an event $\mathscr{F} \in \F$ is denoted by $\Pr(\mathscr{F})$.  For all events $\mathscr{F}, \mathscr{G} \in \F$ with $\Pr(\mathscr{G}) > 0$, the probability of $\mathscr{F}$ conditioned on $\mathscr{G}$ is denoted by $\Pr(\mathscr{F} | \mathscr{G})$.

A random variable is a function with the set of outcomes $\Omega$ as its domain.  For all random variable $Z$, $\Pr(Z \in \mathcal{Z})$ denotes the probability of the event that $Z$ lies within a set $\mathcal{Z}$.  The probability $\Pr(\mathscr{F} | Z = z)$ is of the event $\mathscr{F}$ conditioned on the event $Z = z$.  When $Z$ takes values in $\Re^K$ and has a density $p_Z$, though $\Pr(Z=z)=0$ for all $z$, conditional probabilities $\Pr(\mathscr{F}| Z=z)$ are well-defined and denoted by $\Pr(\mathscr{F} | Z = z)$.
For fixed $\mathscr{F}$, this is a function of $z$.  We denote the value, evaluated at $z=Z$, by $\Pr(\mathscr{F} | Z)$, which is itself a random variable.  Even when $\Pr(\mathscr{F} | Z = z)$ is ill-defined for some $z$, $\Pr(\mathscr{F} | Z)$ is well-defined because problematic events occur with zero probability.  

For each possible realization $z$, the probability $\Pr(Z=z)$ that $Z = z$ is a function of $z$.  We denote the value of this function evaluated at $Z$ by $\Pr(Z)$.  Note that $\Pr(Z)$ is itself a random variable because is it depends on $Z$.  For random variables $Y$ and $Z$ and possible realizations $y$ and $z$, the probability $\Pr(Y=y|Z=z)$ that $Y=y$ conditioned on $Z=z$ is a function of $(y, z)$.  Evaluating this function at $(Y,Z)$ yields a random variable, which we denote by $\Pr(Y|Z)$.

Particular random variables appear routinely throughout the paper.  One is the environment $\environment = (\actions, \observations, \rho)$.  While $\actions$ and $\observations$ are deterministic sets that define the agent-environment interface, the observation probability function $\rho$ is a random variable.  This randomness reflects the agent designer's epistemic uncertainty about the environment.  We often consider probabilities $\Pr(\mathscr{F}|\environment)$ of events $\mathscr{F}$ conditioned on the environment $\environment$.

A policy $\pi$ assigns a probability $\pi(a|h)$ to each action $a$ for each history $h$.  For each policy $\pi$, random variables $A_0^\pi, O_1^\pi, A_1^\pi, O_2^\pi, \ldots$, represent a sequence of interactions generated by selecting actions according to $\pi$.  In particular, with $H_t^\pi = (A_0^\pi, O_1^\pi, \ldots, O_t^\pi)$ denoting the history of interactions through time $t$, we have $\Pr(A^\pi_t|H^\pi_t) = \pi(A^\pi_t|H^\pi_t)$ and $\Pr(O^\pi_{t+1}|H^\pi_t, A^\pi_t, \environment) = \rho(O^\pi_{t+1}|H^\pi_t,A^\pi_t)$.  As shorthand, we generally suppress the superscript $\pi$ and instead indicate the policy through a subscript of $\Pr$.  For example,
$$\Pr_\pi(A_t|H_t) =  \Pr(A^\pi_t|H^\pi_t) = \pi(A^\pi_t|H^\pi_t),$$
and
$$\Pr_\pi(O_{t+1}|H_t, A_t, \environment) =  \Pr(O^\pi_{t+1}|H^\pi_t, A^\pi_t, \environment) = \rho(O^\pi_{t+1}|H^\pi_t, A^\pi_t).$$
The dependence on $\pi$ extends to algorithmic state $Z_t^\pi$, aleatoric state $S_t^\pi$, and epistemic state $P_t^\pi$, and we use the same conventions to suppress superscripts when appropriate.


When expressing expectations, we use the same subscripting notation as with probabilities.  For example, the expectation of a reward $R^\pi_{t+1} = r(S^\pi_t, A^\pi_t, O^\pi_{t+1})$ conditioned on the environment $\environment$, state $S^\pi_t$, and action $A^\pi_t$ is written as $\E[R^\pi_{t+1} |\environment, S^\pi_t, A^\pi_t] = \E_\pi[R_{t+1} | \environment, S_t, A_t]$.

Much of the paper studies properties of interactions under a specific policy $\pi_{\rm agent}$.  When it is clear from context, we suppress superscripts and subscripts that indicate this.  For example, $H_t = H^{\pi_{\rm agent}}_t$, $A_t = A_t^{\pi_{\rm agent}}$, $O_{t+1} = O_{t+1}^{\pi_{\rm agent}}$.  Further,
$$\Pr(A_t|H_t) =  \Pr_{\pi_{\rm agent}}(A_t|H_t) = \pi_{\rm agent}(A_t|H_t).$$

\section{Service Rate Control Example}
\label{se:service-station-appendix}

This appendix supplements the discussion of Section \ref{se:service-station}.  In particular, we provide a precise characterization of environment dynamics and establish that the two baseline agents described in Section \ref{se:service-station} do not deviate from the slow mode of service.  We also present a third, more sophisticated, baseline agent and establish that even that does not learn to deviate from the slow mode.

\subsection{Environment Dynamics}

\begin{figure}[t]
 \centering
 \includegraphics[width=0.6\textwidth]{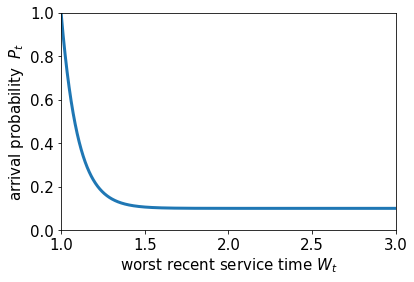}
 \caption{The customer arrival probability is a decreasing function of the maximum service time experienced among the most recent dozen customers served.}\label{fig:arrival_prob}
 \end{figure}

The service station is initially vacant, and customers may arrive starting at the end of the first timestep.  At each time, the arrival probability depends on maximum service time experienced among the most recent 12 customers served.  We denote this statistic by $W_t$ and initialize with $W_0 = 1$.  The customer arrival probability decreases as $W_t$ increases, as illustrated in Figure \ref{fig:arrival_prob}.  In particular, conditioned on $W_t$, the probability that a customer arrives at time $t$, if the service station is vacant then, is $P_t = 0.1 + 0.9 e^{- 10 (W_t - 1)}$.

The choice of service mode impacts service times: with the fast mode, service is always completed in a single timestep, while with the slow mode, the service is completed over the next timestep with probability $1/2$.  As such, the observation probabilities conditioned on $S_t = 0$ are given by

\begin{centering}

\begin{tabular}{|c|cc|}
\hline
$\rho_*(o|H_t, \cdot)$ &  $\mathrm{departure}$ & $\neg\mathrm{departure}$ \\
\hline
$\mathrm{arrival}$ & $0$ & $P_t$ \\
$\neg\mathrm{arrival}$ & $0$ & $1-P_t$ \\
\hline
\end{tabular}

\end{centering}

and, conditioned on $S_t = 1$,

\begin{centering}

\begin{tabular}{|c|cc|}
\hline
$\rho_*(o|H_t, \mathrm{fast})$ &  $\mathrm{departure}$ & $\neg\mathrm{departure}$ \\
\hline
$\mathrm{arrival}$ & $P_t$ & $0$ \\
$\neg\mathrm{arrival}$ & $1-P_t$ & $0$ \\
\hline
\end{tabular}
\begin{tabular}{|c|cc|}
\hline
$\rho_*(o|H_t, \mathrm{slow})$ &  $\mathrm{departure}$ & $\neg\mathrm{departure}$ \\
\hline
$\mathrm{arrival}$ & $P_t/2$ & $0$ \\
$\neg\mathrm{arrival}$ & $(1-P_t)/2$ & $1/2$ \\
\hline
\end{tabular}

\end{centering}

It is easy to verify that the long-run average reward is maximized if the agent applies the fast mode of service over every timestep.  This policy minimizes service times, with each customer waiting for precisely one timestep.  Consequently, under this policy, $W_t$ converges to $1$, as does the arrival probability.  The long-run average reward is therefore $1-0.5= 0.5$. 

\subsection{Analysis of Baseline Agents}

We now study the performance of the two baseline agents introduced in Section \ref{se:service-station}, as well a more sophisticated variant.  Recall that these agents at each time apply a policy $\pi_\epsilon$, which selects the fast mode with some probability $\epsilon$, which can vary with time.  Each of these agents begins with knowledge of service completion probabilities: $1/2$ and $1$ for the slow and fast modes, respectively.
Throughout the discussion,  Let $W_\infty$ and $P_\infty$ denote random variables sampled from the steady-state distributions of $W_t$ and $P_t$, respectively.  

Let $c_\epsilon$ be the service completion probability over any timestep when a customer is served under policy $\pi_\epsilon$.  In particular, $c_{\epsilon} =\epsilon + \frac{1}{2}(1-\epsilon) = \frac{1+\epsilon}{2}$.  The average reward $\lambda_{\pi_\epsilon}$ is then given by
\begin{equation}\label{eq:average_reward_starting_point}
\lambda_{\pi_\epsilon} 
    =  \frac{\text{profit per customer}}{\text{mean interarrival time}} = \frac{1 - 0.5 \frac{\epsilon}{c_{\epsilon}}}{\frac{1}{c_{\epsilon}} -1 + \frac{1}{\E_{\pi_{\epsilon}}[P_\infty \mid \environment = e_*]}}.
\end{equation}

The first agent applies $\pi_0$ and only deviates if warranted after observing data over a long duration, assuming that the arrival probability is fixed.  Whether it decides to increase $\epsilon$ depends on its arrival probability estimate.  Under the policy $\pi_0$, a customer's service time is $1$ with probability $1/2$ and, otherwise, at least $2$. Since $W_t$ is the largest among $12$ service times, $\Pr_{\pi_0}(W_\infty = 1 \mid \environment = e_*) = 1/4096$ and $\Pr_{\pi_0}(W_\infty \geq 2 \mid \environment = e_*) = 4095/4096$.  Consequently, 
\begin{align*}
\E_{\pi_0}\left[P_\infty \mid \environment = e_*\right] = \E_{\pi_0}\left[0.1 + 0.9 e^{- 10 (W_\infty - 1)}\mid \environment = e_* \right] < 1.
\end{align*}

To keep things simple, suppose the agent's estimate of this steady-state arrival probability is exactly $\E_{\pi_0}\left[P_\infty \mid \environment = e_*\right]$. 
As such, the agent's \textit{estimate} of the average reward under $\pi_\epsilon$ is
\begin{equation}
\label{eq:average-reward-estimate-formula}
\hat{\lambda}_{\pi_\epsilon} = \frac{\text{profit per customer}}{\text{mean interarrival time}} = \frac{1 - 0.5\epsilon \frac{1}{c_\epsilon}}{\frac{1}{c_\epsilon} - 1 + \frac{1}{\E_{\pi_0}[P_\infty \mid \environment = e_*]}} = \frac{1}{1 - \epsilon + \frac{1+\epsilon}{\E_{\pi_0}[P_\infty \mid \environment = e_*]}}.
\end{equation}
Note that the difference between Equation~\eqref{eq:average-reward-estimate-formula} and Equation~\eqref{eq:average_reward_starting_point} is due to the first agent's use of an arrival probability that results from $\pi_0$ rather than $\pi_{\epsilon}$.
Since $\E_{\pi_0}\left[P_\infty \mid \environment = e_* \right] < 1$, $\hat{\lambda}_{\pi_\epsilon}$ is strictly decreasing in $\epsilon \in [0,1]$.
As such, the agent does not deviate from the slow mode.

The second agent additionally tries a small value of $\epsilon > 0$ for some duration in order to estimate the derivative $\mathrm{d} \lambda_{\pi_\epsilon} / \mathrm{d}\epsilon$ of the average reward at $\epsilon = 0$. If this derivative is positive, it increases $\epsilon$. Here, we show
that this derivative is negative, and hence the second agent does not deviate from the slow mode.
As such, one can easily check that for each positive integer $w$ and for all $\epsilon \in [0,1]$,
we have:
\begin{equation}\label{eq:prob_W_inf}
\Pr_{\pi_{\epsilon}}(W_\infty \le w \mid \environment = e_* )
= \left(1 - \left(\frac{1-\epsilon}{2}\right)^w\right)^{12}.
\end{equation}


Recalling that $c_{\epsilon} = \frac{1+\epsilon}{2}$ and following Equation~\eqref{eq:average_reward_starting_point},
we have:
\begin{align}
    \lambda_{\pi_\epsilon}  =
    \frac{1 - 0.5\epsilon \frac{1}{c_{\epsilon}}}{\frac{1}{c_{\epsilon}} - 1 + \frac{1}{\E_{\pi_{\epsilon}}[P_{\infty} \mid \environment = e_* ]}} 
    = \frac{G(\epsilon)}{(1 - \epsilon)G(\epsilon) + (1+\epsilon) },\label{eq:average_reward1}
\end{align}
where $G(\epsilon) = \E_{\pi_\epsilon}[P_\infty \mid \environment = e_* ] = \sum_{w=1}^\infty \Pr_{\pi_{\epsilon}}(W_\infty = w \mid \environment = e_*) \left(0.1 + 0.9 e^{-10(w-1)}\right)$.
Consequently, taking its derivative with respect to $\epsilon$ and evaluating it $\epsilon = 0$ yields:
\begin{equation}\label{eq:average_reward_second_agent}
\frac{\mathrm{d} \lambda_{\pi_\epsilon}}{\mathrm{d}\epsilon}\Bigg|_{\epsilon = 0}
= \frac{\frac{\mathrm{d}}{\mathrm{d}\epsilon}G(\epsilon) |_{\epsilon=0} - G(0) + G(0)^2}{\big[ G(0) + 1\big]^2} < -0.072.
\end{equation}
As such, the second agent will not deviate from $\pi_0$.

Finally, we can even consider a third agent, which is similar to the second agent except it additionally estimates the second derivative $\mathrm{d}^2 \lambda_{\pi_\epsilon} / \mathrm{d}\epsilon^2$.  It then chooses $\epsilon$ to maximize a second-order Taylor expansion of $\lambda_{\pi_\epsilon}$ around $\epsilon=0$ subject to the constraint $0\leq \epsilon \leq 1$. This agent again ends up always selecting the slow mode of service, because even exploiting second-order information suggests that staying with $\pi_0$ is the best thing to do. To see this, 
we evaluate the second derivative of $\lambda_{\pi_\epsilon}$ at $\epsilon = 0$, yielding $\frac{1}{2}\frac{\mathrm{d}^2 \lambda_{\pi_\epsilon}}{\mathrm{d} \epsilon^2}\Big|_{\epsilon = 0} < 0.0716$.
As such, when using a second-order polynomial for extrapolation, one would 
get $\tilde{\lambda}_{\pi_\epsilon} = \lambda_{\pi_0} - \frac{\mathrm{d} \lambda_{\pi_\epsilon}}{\mathrm{d} \epsilon}\Big|_{\epsilon = 0}\epsilon +  \frac{1}{2}\frac{\mathrm{d}^2 \lambda_{\pi_\epsilon}}{\mathrm{d} \epsilon^2}\Big|_{\epsilon = 0}\epsilon^2
 < \lambda_{\pi_0} - 0.072\epsilon +  0.0716\epsilon^2 < 0
$, thereby yielding a strictly smaller value than $\lambda_{\pi_0}$ for all $\epsilon \in (0,1]$.

In summary, the first baseline agent represents what might be produced by a conservative designer, who demands to see empirical evidence justifying fast service before ever trying that.  The second agent is representative of approaches used in the policy gradient literature, as discussed in \citep{sutton2018reinforcement} and references therein.  The third agent pursues a more sophisticated approach entailing estimation and use of the second derivative in addition to the gradient. Per the analysis given above, all three agents end up choosing the slow-only policy,  and hence perform poorly relative to our optimistic Q-learning agent, which adapts action values $Q_t(S_t,A_t)$ to predict future return and select actions.

\section{Proofs}

\subsection{Proof of Lemma \ref{lemma:mixing-time}}
\label{subsec:proof-of-lemma-mixing-time}

The lemma is restated below.

\mixingtimelemma*

To prove this lemma, recall that, for a fixed policy $\pi\in\policies$, discount factor $\gamma\in[0,1)$, and history $h\in\histories$, 
\[
   V_\pi^\gamma(h) = \sum_{t=0}^\infty \Big(\gamma^t\cdot \big(P_\pi^t \overline{r}_\pi\big) (h)\Big).
\]
For simplicity, let $r_\ell = \gamma^{\ell}\cdot \big(P_\pi^\ell \overline{r}_\pi\big) (h)$.  We have that
\[
    V_\pi^\gamma(h) = \sum_{\ell=0}^\infty \gamma^\ell r_\ell.
\]
By the definition of $\tau_\pi$, for all $\ell \geq 0$,
\[
    \tau_\pi \geq \left| \sum_{k=0}^\ell \big(r_k - \lambda_\pi\big)\right|.
\]
Hence,
\begin{eqnarray}
    \left| V_\pi^\gamma(h) - \frac{\lambda_\pi}{1 - \gamma}\right|
    &=& \left| \sum_{\ell=0}^\infty \gamma^\ell\cdot\big(r_\ell - \lambda_\pi\big)\right|\nn\\
    &=& \left|\sum_{\ell=0}^\infty(1-\gamma)\gamma^\ell \cdot \sum_{k=0}^\ell \big(r_k - \lambda_\pi\big) \right|\nn\\
    &\leq& \sum_{\ell=0}^\infty(1-\gamma)\gamma^\ell \cdot\left|\sum_{k=0}^\ell \big(r_k - \lambda_\pi\big) \right|\nn\\
    &\leq& \sum_{\ell=0}^\infty(1-\gamma)\gamma^\ell \cdot \tau_\pi\nn\\
    &=& \tau_\pi,\nn
\end{eqnarray}
which is our desired result.

\subsection{Properties of the Learning Rates}
\label{subsec:properties-of-stepsize}
In this subsection, we generalize a useful lemma from \citep{jin2018q} on properties of the learning rates.
Let 
\[
    \alpha_k^i = \alpha_i \cdot \prod_{\ell=i+1}^k ( 1- \alpha_\ell), \quad i=1,\dots,k,
\]
where $(\alpha_\ell:\ell = 1,2,\dots)$ is the learning rate sequence $\alpha_\ell = (1+2\tau)/(\ell+2\tau)$, 
and
\[
    \alpha_k^0 = {\bf 1}\{k=0\}.
\]
Naturally, $\sum_{i=0}^k \alpha_k^i = 1$.  We also have the following:
\begin{lemma}
\label{lemma:learning-rate} $ $
\begin{enumerate}[label=(\alph*)]
    \item For all $k\geq 1$, $\frac{1}{\sqrt{k}} \leq \sum_{i=1}^k \frac{\alpha_k^i}{\sqrt{i}} \leq \frac{2}{\sqrt{k}}$;
    \item For all $k\geq 1$, $\max_{i=1,\dots,k}\alpha_k^i \leq \frac{4\tau}{k}$ and $\sum_{i=1}^k (\alpha_k^i)^2 \leq \frac{4\tau}{k}$;
    \item For all $i\geq 1$, $\sum_{k=i}^\infty \alpha_k^i = 1 + \frac{1}{2\tau}$.
\end{enumerate}
\end{lemma}
\noindentproof 
The proof of Lemma 4.1 in \citep{jin2018q} covers the case where $H=2\tau$ is a positive integer.  We note that their proof of parts $(a)$ and $(b)$ also applies for all $H>1$.  Thus, what is left for us here is showing that part $(c)$ holds for all real numbers $H>1$.  To this end, we first establish that, for all positive real numbers $b>a$,
\begin{equation}
    \label{eq:learning-rate-lemma-1}
    \frac{a}{b-a} = \sum_{i=1}^\infty\prod_{j=1}^i \frac{a+j-1}{b+j}.
\end{equation}

In fact, we can show by induction on positive integer $\ell$ that
\begin{equation}
    \label{eq:learning-rate-lemma-2}
    \frac{a}{b-a} = \sum_{i=1}^\ell \prod_{j=1}^i \frac{a+j-1}{b+j} + \frac{a}{b-a}\prod_{j=1}^\ell \frac{a+j}{b+j}.
\end{equation}
When $\ell = 1$, we have that
\[
    \frac{a}{b-a} - \frac{a}{b+1} = \frac{a}{b-a}\cdot \left[1 - \frac{b-a}{b+1} \right] = \frac{a}{b-a}\cdot\frac{a+1}{b+1}.
\]
Hence, \eqref{eq:learning-rate-lemma-2} holds for $\ell = 1$.  Now suppose that \eqref{eq:learning-rate-lemma-2} holds for $\ell$.  There is
\begin{eqnarray}
    \frac{a}{b-a} - \sum_{i=1}^{\ell+1} \prod_{j=1}^i \frac{a+j-1}{b+j}
    &=& \left\{\frac{a}{b-a} - \sum_{i=1}^{\ell} \prod_{j=1}^i \frac{a+j-1}{b+j}\right\} - \prod_{j=1}^{\ell+1} \frac{a+j-1}{b+j}\nn\\
    &=& \frac{a}{b-a}\prod_{j=1}^\ell \frac{a+j}{b+j}  - \prod_{j=1}^{\ell+1} \frac{a+j-1}{b+j}\label{eq:learning-rate-lemma-3}\\
    &=& \left\{\frac{a}{b-a} \prod_{j=1}^\ell \frac{a+j}{b+j}\right\}\cdot \left(1 - \frac{b-a}{b+\ell+1}  \right) \nn\\
    &=& \frac{a}{b-a} \prod_{j=1}^{\ell+1} \frac{a+j}{b+j},
\end{eqnarray}
where \eqref{eq:learning-rate-lemma-3} follows from our induction hypothesis.
Thus, \eqref{eq:learning-rate-lemma-2} also holds for $\ell+1$, concluding our induction.  Following \eqref{eq:learning-rate-lemma-2}, we have
\begin{equation}
    \label{eq:learning-rate-lemma-4}
    \frac{a}{b-a} - \sum_{i=1}^\infty \prod_{j=1}^i \frac{a+j-1}{b+j} 
    = \lim_{\ell\to\infty}\left\{\frac{a}{b-a} - \sum_{i=1}^\infty \prod_{j=1}^i \frac{a+j-1}{b+j} \right\}
    = \lim_{\ell\to\infty}\frac{a}{b-a}\prod_{j=1}^\ell \frac{a+j}{b+j}.
\end{equation}
However,
\[
    \log \prod_{j=1}^\ell \frac{a+j}{b+j} = \sum_{j=1}^\ell \log\left( 1 - \frac{b-a}{b+j}\right) \leq -\sum_{j=1}^\ell  \frac{b-a}{b+j}.
\]
The right-hand side goes to $-\infty$ as $\ell\to\infty$, implying that 
\[
    \lim_{\ell\to\infty}\prod_{j=1}^\ell \frac{a+j}{b+j} = 0.
\]
Thus, \eqref{eq:learning-rate-lemma-1} follows from \eqref{eq:learning-rate-lemma-4}.  Now we have
\begin{eqnarray}
    \sum_{k=i}^\infty \alpha_k^i
    &=& \frac{H+1}{H+i} \cdot \left\{ 1 + \sum_{k=i}^\infty\prod_{j=0}^{k-i}\frac{i+j}{H+i+j+1} \right\} \nn\\
    &=& \frac{H+1}{H+i} \cdot \left\{ 1 + \sum_{k=1}^\infty\prod_{j=1}^{k}\frac{i+j-1}{H+i+j} \right\} \nn\\
    &=& \frac{H+1}{H+i} \cdot \left\{ 1 + \frac{i}{H}\right\}\nn\\
    &=& \frac{H+1}{H}\nn\\
    &=& 1 + \frac{1}{2\tau},
\end{eqnarray}
as we have claimed in part $(c)$. \qedsymbol

\subsection{Regret Analysis of the Discounted $Q$-Learning Agent}
\label{sec:regret-analysis}

In this section, we focus ourselves on the discounted variant of the agent in Section \ref{subsubsec:discounted-q-learning}.  Throughout this section we assume that the discount factor $\gamma = 1 - 1/\tau \in[0,1)$ is fixed.  We also consider a hypothetical setting, in which the history starts from an arbitrary $h\in\histories$ and the agent starts from aleatoric state $\phi(h)$.  Since every time the agent changes the discount factor, the environment history is not reset to $\emptyH$, our main goal here is to demonstrate that our result holds regardless of the initial history, as long as the agent starts from the corresponding agent state.  In order to simplify notations, in this section we always have $H_0=h$, which is a fixed, possibly non-empty history, and $S_0 = \phi(h)$.  We will also omit the superscript $\gamma$ on value functions $V_*^\gamma$, $Q_*^\gamma$, $V_\pi^\gamma$ and $Q_\pi^\gamma$.  Readers should keep in mind that all value functions in this section are with respect to discount factor $\gamma$.

\begin{algorithm}
    \begin{algorithmic}[1]
    \STATE{\bf Input: } $\falea, r, T, \gamma,\beta, Q_{\rm init}$
    \STATE initialize history to $h$
    \STATE $t = 0,\quad s \leftarrow \phi(h)$ 
    \STATE $Q\leftarrow Q_{\rm init}, \quad N(\cdot, \cdot) \leftarrow 0$
    \STATE $V(s) \leftarrow \max_{a'\in\actions} Q(s,a'),\quad\forall s\in\states$
    \WHILE{$t < T$}
        \STATE $a \leftarrow \mathtt{sample\_unif}(\argmax_{a' \in \actions} Q(s,a'))$ 
        \STATE $N(s,a) \leftarrow N(s,a) + 1$
        \STATE $\alpha \leftarrow \frac{2+ (1-\gamma)}{2 + N(s,a)\cdot(1 - \gamma)}$
        \STATE execute action $a$ and register observation $o$
        \STATE $s' \leftarrow \falea(s,a,o)$ 
        \STATE $Q(s,a) \leftarrow (1-\alpha)\cdot Q(s,a) + \alpha \cdot \Big[ r(s,a,o) + \gamma \cdot  V(s') + \frac{\beta}{\sqrt{N(s,a)}}\Big]$ 
        \STATE $V(s) \leftarrow \min\Big\{\max_{a'\in\actions} Q(s,a'),\ 1/(1-\gamma) \Big\}$
        \STATE $s \leftarrow s',\quad t \leftarrow t+1$
    \ENDWHILE
    \end{algorithmic}
    \caption{Discounted $Q$-learning subroutine}
    \label{alg:fictional-q-learning-discounted}
\end{algorithm}\par

Specifically, we will consider Algorithm \ref{alg:fictional-q-learning-discounted}, which is identical to Algorithm \ref{alg:q-learning-update} except that the initial history can be arbitrary, and that we use $Q_{\rm init}(s,a)$ to initialize $Q(s,a)$ for all $s\in\states$ and $a\in\actions$. Let $H_1,H_2,\dots$ be the history trajectory of Algorithm \ref{alg:fictional-q-learning-discounted}, i.e.
\[
    H_t = \big(h, A_0, O_1, \dots, A_{t-1}, O_t\big),\quad t= 1,2, \dots,
\]
and let 
\[
    S_t = \phi(H_t),\quad R_{t} = r(S_{t-1}, A_{t-1}, O_{t}),\quad t=1,2, \dots.
\]
Also let $V_t(s)$ be the value of aleatoric state $s$ at timestep $t$ immediately {\it after} the update 
\[
    Q(S_{t-1}, A_{t-1}) \leftarrow (1 - \alpha) \cdot Q(S_{t-1}, A_{t-1}) + \alpha\cdot \left(R_t + \gamma\cdot V(S_t) + \frac{\beta}{\sqrt{N(S_{t-1}, A_{t-1})}}\right)
\]
and
\[
    V(S_{t-1}) \leftarrow \min\Bigg\{\max_{a'\in\actions} Q(S_{t-1},a'),\ \frac{1}{1-\gamma} \Bigg\}.
\]
and let $V_t(h)$ be a shorthand for $V_t(\phi(h))$. Similarly are $Q_t(s,a)$ and $Q_t(h,a)$ defined.
Note that since the actions are selected greedily, we have
\[
    \label{eq:appendix-13}
    Q_t(H_t, A_t) = \max_{a\in\actions} Q_t(H_t, a) = V_t(H_t).
\]
Finally, we let $\trans$ be the {\it transition operator}, such that for all functions $g:\histories\mapsto\mathbb{R}$ and history-action pairs $(h,a)\in\histories\times\actions$, 
\[
    \trans g (h,a) = \sum_{h'\in\histories} \Big(P_{ahh'}\cdot g(h') \Big).
\]

Let $\hat{\pi}$ be the policy corresponding to Algorithm \ref{alg:fictional-q-learning-discounted}.  
Recall that the distortion with respect to effective planning horizon $\tau\geq 1$ is defined as
\[
    \Delta_\tau = \max_{(s,a) \in \states\times\actions} \left(\sup_{h \in \histories: \phi(h) = s} Q_*^{1-1/\tau}(h,a) - \inf_{h \in \histories: \phi(h) = s} Q_*^{1-1/\tau}(h,a)\right).
\]
To avoid cluttering, for $\gamma\in[0,1)$ we will simply use $\Delta_\gamma$ to represent $\Delta_\tau$ with $\tau = 1/(1-\gamma)$.
Our aim is to show the following result:
\begin{theorem}
\label{theorem:discounted}
If Algorithm \ref{alg:fictional-q-learning-discounted} is executed with $\gamma\in[0,1)$,
\[
    \beta = \frac{4}{(1-\gamma)^{3/2}}\sqrt{\log (2T^2)},
\]
and $Q_{\rm init}$ such that for some $\iota \geq 0$,
\[
    \label{eq:q-init-condition}
    Q_*^\gamma(h,a) - \frac{\iota}{1-\gamma}\leq Q_{\rm init}\big(\phi(h),a\big) \leq \frac{1}{1-\gamma},\quad \forall h\in\histories, a\in\actions,
\]
then for all $T\geq 1$ and initial history $h\in\histories$,
\[
    \label{eq:theorem-discounted-1}
    \E_{\hat{\pi}} \left[ \sum_{t=0}^{T-1} \bigg(V_*(H_t) - V_{\hat{\pi}} (H_t)\bigg)\right] \leq \frac{24}{(1-\gamma)^{\frac{5}{2}}}\cdot\sqrt{\states\actions T\cdot \log(2T^2)} + \frac{3\tilde{\Delta}_\gamma T}{1-\gamma} + \frac{\states\actions+3}{(1-\gamma)^2},
\]
where $\tilde{\Delta}_\gamma = \max\{\Delta_\gamma, \iota\}$.
\end{theorem}
 
\subsubsection{Regret Decomposition}
In this subsection we will prove the following lemma, which decomposes the left-hand side of \eqref{eq:theorem-discounted-1} and paves the way for further analysis.

\begin{lemma}
    \label{lemma:regret-decomposition}
    For all $T\geq 1$,
    \[
        \E_{\hat{\pi}} \left[ \sum_{t=0}^{T-1} \bigg(V_*(H_t) - V_{\hat{\pi}} (H_t)\bigg)\right] &\leq& \frac{1}{1-\gamma}\cdot \E_{\hat{\pi}} \left[ \sum_{t=0}^{T-1} \bigg(V_*(H_t) - Q_{*} (H_t, A_t)\bigg)\right]\nn\\
            &&+\ \frac{1}{(1-\gamma)^2}.\label{eq:lemma-regret-decomposition-1}
    \]
\end{lemma}
\noindentproof
We have 
\[
    \E_{\hat{\pi}} \left[ \sum_{t=0}^{T-1} \bigg(V_*(H_t) - V_{\hat{\pi}} (H_t)\bigg)\right]
    &=& \E_{\hat{\pi}} \left[ \sum_{t=0}^{T-1} \bigg(V_t(H_t) - V_{\hat{\pi}} (H_t)\bigg)\right] \nn\\
        &&-\ \E_{\hat{\pi}} \left[ \sum_{t=0}^{T-1} \bigg(V_t(H_t) - V_{*} (H_t)\bigg)\right].\label{eq:appendix-25}
\]
Taking a closer look at the first term on the right-hand side, since $\hat{\pi}$ is greedy with respect to $V_t$ for each $t$,
\[
    \E_{\hat{\pi}} \left[ \sum_{t=0}^{T-1} \bigg(V_t(H_t) - V_{\hat{\pi}} (H_t)\bigg)\right]
    &=& \E_{\hat{\pi}} \left[ \sum_{t=0}^{T-1} \bigg(V_t(H_t) - Q_{\hat{\pi}} (H_t, A_t)\bigg)\right]\nn\\
    &=& \E_{\hat{\pi}} \left[ \sum_{t=0}^{T-1} \bigg(V_t(H_t) - Q_{*} (H_t, A_t)\bigg)\right]\nn\\
        &&+\ \E_{\hat{\pi}} \left[ \sum_{t=0}^{T-1} \bigg(Q_*(H_t, A_t) - Q_{\hat{\pi}} (H_t, A_t)\bigg)\right]\nn\\
    &\leq& \E_{\hat{\pi}} \left[ \sum_{t=0}^{T-1} \bigg(V_t(H_t) - Q_{*} (H_t, A_t)\bigg)\right]\nn\\
        &&+\ \gamma\cdot \E_{\hat{\pi}} \left[ \sum_{t=0}^{T-1} \bigg(V_*(H_{t+1}) - V_{\hat{\pi}} (H_{t+1})\bigg)\right].\label{eq:appendix-26}
\]
Combining \eqref{eq:appendix-25} and \eqref{eq:appendix-26}, we have
\[
    (1-\gamma)\cdot  \E_{\hat{\pi}} \left[ \sum_{t=0}^{T-1} \bigg(V_*(H_t) - V_{\hat{\pi}} (H_t)\bigg)\right]
    &\leq& \gamma\cdot \E_{\hat{\pi}} \left[ \sum_{t=0}^{T-1} \bigg(V_*(H_t) - Q_{*} (H_t, A_t)\bigg)\right]\nn\\
        &&+\ \bigg(V_*(H_{T}) - V_{\hat{\pi}} (H_{T})\bigg).
\]
Dividing both sides by $1-\gamma$ and considering that $V_*(H_T) - V_{\hat{\pi}} (H_{T})\leq 1/(1-\gamma)$, we arrive at \eqref{eq:lemma-regret-decomposition-1}.\qedsymbol

For simplicity, let...
\[
    \label{eq:def-chi}
    \chi_k = V_{k}(H_k) - V_*(H_k) + \frac{\tilde{\Delta}_\gamma}{1 - \gamma}
\]
and
\[
    \label{eq:def-xi}
    \xi_k = Q_k(H_k, A_k) - Q_*(H_k, A_k)
\]
for each $k\geq 0$.  Using these notations, \eqref{eq:lemma-regret-decomposition-1} can be written equivalently as
\[
    \E_{\hat{\pi}} \left[ \sum_{t=0}^{T-1} \bigg(V_*(H_t) - V_{\hat{\pi}} (H_t)\bigg)\right] \leq 
    \frac{1}{1-\gamma}\cdot \E_{\hat{\pi}} \left[ \sum_{t=0}^{T-1} \big(\xi_k - \chi_k\big)\right] + \frac{\tilde{\Delta}_\gamma T}{(1-\gamma)^2} + \frac{1}{(1-\gamma)^2}.
    \label{eq:after-decomposition-1}
\]

\subsubsection{Establishing Near-Optimism}
\label{subsec:optimism}

In this subsection we show that at each timestep $t$, the value function $V_t$ is almost optimistic uniformly across all histories.  
We have the following result.

\begin{lemma}
\label{lemma:optimism}
If Algorithm \ref{alg:fictional-q-learning-discounted} is executed with $\gamma\in[0,1)$,
\[
    \beta_\delta=\frac{4}{(1 - \gamma)^{3/2}}\sqrt{\log \frac{2T}{\delta}},
\]
and $Q_{\rm init}$ such that for some $\iota \geq 0$,
\[
    Q_*^\gamma(h,a) - \frac{\iota}{1-\gamma}\leq Q_{\rm init}(h,a) \leq \frac{1}{1-\gamma},\quad \forall h\in\histories, a\in\actions,
\]
then with probability at least $1-\delta$, for all $h\in\histories, a\in\actions$ and $0\leq t\leq T$,
$$
    V_t(h) \geq V_*(h) - \frac{\tilde{\Delta}_\gamma}{1-\gamma}\quad\text{and}\quad 
    Q_t(h,a) \geq Q_*(h,a) - \frac{\tilde{\Delta}_\gamma}{1-\gamma},
$$
where $\tilde{\Delta}_\gamma = \max\{\Delta_\gamma, \iota\}$.
\end{lemma}
\noindentproof
For the moment let us fix $h\in\histories$ and $a\in\actions$.  Let $\hat{Q}_k$ be the $Q$-value of $\big(\phi(h), a\big)$ after it has been updated $k$ times, with $\hat{Q}_0 = 1/(1-\gamma)$ being the initial value.  Further, for each $k=1,2,\dots$, let $t_k$ be the timestep at which $\big(\phi(h), a\big)$ is updated.  Note that we have
\[
    \label{eq:appendix-12}
    \phi(H_{t_k}) = \phi(h),\quad A_{t_k} = a.
\]
From the update rule in Algorithm \ref{alg:fictional-q-learning-discounted}, when $n\geq 1$,
\[
    \hat{Q}_n = \sum_{i=1}^{n} \alpha_{n}^i \cdot \left(
        R_{t_i+1} + \gamma\cdot V_{t_i}\big( H_{t_i+1}\big) + \frac{\beta_\delta}{\sqrt{i}}
    \right).
\]

Thus, when $n\geq 1$,
\begin{eqnarray}
    \hat{Q}_n - Q_*(h, a)
    &=& 
    \sum_{i=1}^{n} \alpha_{n}^i \cdot \left(
        R_{t_i+1} + \gamma\cdot V_{t_i}\big( H_{t_i+1}\big) + \frac{\beta_\delta}{\sqrt{i}}
        - Q_*(h, a)
    \right)\nn\\
    &\geq& 
    \sum_{i=1}^{n} \alpha_{n}^i \cdot \Bigg(
        R_{t_i+1} + \gamma\cdot V_{t_i}\big( H_{t_i+1}\big) + \frac{\beta_\delta}{\sqrt{i}}\Bigg)\nn\\
        && - \sum_{i=1}^{n} \alpha_{n}^i \cdot \Bigg(Q_*(H_{t_i}, A_{t_i}) + \tilde{\Delta}_\gamma
    \Bigg)\label{eq:appendix-1}\\
    &=& 
    \sum_{i=1}^{n} \alpha_{n}^i \cdot \Bigg(
        \gamma\cdot V_{t_i}\big( H_{t_i+1}\big) - \gamma\cdot\trans V_*(H_{t_i}, A_{t_i}) \Bigg)\nn\\ && +\ \Bigg\{\sum_{i=1}^{n} \alpha_{n}^i \cdot \frac{\beta_\delta}{\sqrt{i}} 
    \Bigg\} - \tilde{\Delta}_\gamma \label{eq:appendix-2}\\
    &\geq& 
    \sum_{i=1}^{n} \alpha_{n}^i \cdot \Big(
         V_{t_i}( H_{t_i+1} ) - V_*(H_{t_i+1})
        \Big) \nn\\
        &&+\ 
        \sum_{i=1}^{n} \alpha_{n}^i \cdot \Big(
        V_*(H_{t_i+1}) - \trans V_*(H_{t_i}, A_{t_i})
    \Big) + \frac{\beta_\delta}{\sqrt{n}} - \tilde{\Delta}_\gamma,\label{eq:optimism-discounted}
\end{eqnarray}
where \eqref{eq:appendix-1} follows from \eqref{eq:appendix-12} and that
\[
    \phi(H_{t_i}) = \phi(h) \quad \Rightarrow\quad \left|Q_*(h,a) - Q_*(H_{t_i}, a) \right| \leq \Delta_\gamma\leq\tilde{\Delta}_\gamma,\ \forall a\in\actions;
\]
and \eqref{eq:appendix-2} follows from the fact that
\[
    Q_*(H_{t_i}, A_{t_i}) = R_{t_i+1} + \gamma\cdot \trans V_*(H_{t_i}, A_{t_i}).
\]

Consider the following sequence:
\[
    G_k = \sum_{i=1}^k \alpha_{n}^i \cdot \bigg(
        V_*(H_{t_i+1}) - \trans V_*(H_{t_i}, A_{t_i})
    \bigg),\quad k=1,\dots, n,
\]
with $G_0 = 0$.
We have that, for $k\geq 1$,
\[
    \E[G_k | G_{k-1}] 
    &=& \E\left[ 
        \alpha_{n}^k \cdot \bigg(
        V_*(H_{t_k+1}) - \trans V_*(H_{t_k}, A_{t_k})
    \bigg)\right]\nn\\
    &=& \E\left[ \E\left[
        \alpha_{n}^k \cdot \bigg(
        V_*(H_{t_k+1}) - \trans V_*(H_{t_k}, A_{t_k})
    \bigg)\bigg| H_{t_k}, A_{t_k}\right]\right]\nn\\
    &=& 0,
\]
implying that $\{G_k:k=0,\dots, n\}$ is a martingale.  As a result, it follows from Azuma-Hoeffding inequality that, with probability at least $1-\delta$,
\[
    \label{eq:azuma-hoeffding}
    \big|G_{n} - G_0\big|
    &=& 
    \left|
    \sum_{i=1}^{n} \alpha_{n}^i \bigg(
        V_*(H_{t_i+1}) - \trans  V_*(H_{t_i}, A_{t_i})
    \bigg)
    \right|\nn\\
    &\leq& \frac{4}{(1-\gamma)^{3/2}} \cdot \frac{1}{\sqrt{n}} \cdot \sqrt{\log\frac{2}{\delta}},
\]
where we used assertion $(b)$ of Lemma \ref{lemma:learning-rate} and the fact that
\[
    \Big|
        V_*(H_{t_i+1}) - \trans  V_*(H_{t_i}, A_{t_i})
    \Big| \leq \frac{1}{1 - \gamma}.
\]

Scaling $\delta$ to $\delta/T$ and applying union bounds, we have that, with probability at least $1-\delta$, simultaneously for all $h\in\histories, a\in\actions$ and $n\geq 1$, as long as $\big(s(h), a\big)$ is updated not more than $n$ times in timesteps $1,2\dots, T$, 
\[
    \left|
    \sum_{i=1}^{n} \alpha_{n}^i \bigg(
        V_*(H_{t_i+1}) - \trans  V_*(H_{t_i}, A_{t_i})
    \bigg)
    \right|
    \leq \frac{4}{(1-\gamma)^{3/2}} \cdot \frac{1}{\sqrt{n}} \cdot \sqrt{\log\frac{2T}{\delta}}.
    \label{eq:high-probability-event}
\]
We denote the above event by $\mathfrak{E}$.
Recall that we choose
\[
    \beta_\delta = \frac{4}{(1-\gamma)^{3/2}} \cdot \sqrt{\log\frac{2T}{\delta}}.
\]

As a result, following \eqref{eq:optimism-discounted}, conditioned on event $\mathfrak{E}$,
\[
    \label{eq:hat-Q}
    \hat{Q}_n - Q_* \geq \gamma \cdot \sum_{i=1}^{n} \alpha_{n}^i \cdot \left(
        V_{t_i}\big( H_{t_i+1}\big) - V_*(H_{t_i+1})
        \right) - \tilde{\Delta}_\gamma.
\]
We will now show our desired result by induction. Assume that event $\mathfrak{E}$ occurs. At $t=0$, from our requirements on $Q_{\rm init}$, obviously there is $V_0(h) \geq V_*(h) -\tilde{\Delta}_\gamma/(1-\gamma)$ for all $h\in\histories$.  Suppose that the result holds for all $t<t'$.  At $t=t'$, for all $(h,a)\in\states\times\actions$, as long as $\big(\phi(h), a\big)$ is updated $n\geq 1$ times in timesteps $1,2\dots, t'$, from \eqref{eq:hat-Q} we have
\begin{eqnarray}
    Q_{t'}(h,a) - Q_*(h,a)
    &\geq& \gamma \cdot \sum_{i=1}^{n} \alpha_{n}^i \cdot \left(
        V_{t_i}\big( H_{t_i+1}\big) - V_*(H_{t_i+1})
    \right) - \tilde{\Delta}_\gamma\nn\\
    &\geq& \gamma\cdot\left(-\frac{\tilde{\Delta}_\gamma}{1-\gamma}\right) - \tilde{\Delta}_\gamma \nn\\
    &=& -\frac{\tilde{\Delta}_\gamma}{1 - \gamma}.
\end{eqnarray}
Otherwise, if $\big(\phi(h), a\big)$ is not updated in timesteps $1,2,\dots,t'$, then $V_{t'}(h) = V_0(h) \geq V_*(h) -\tilde{\Delta}_\gamma/(1-\gamma)$.
This leads to 
$$
    V_{t'}(h)- V_*(h) 
    = 
    \min\Bigg\{\max_{a'\in\actions}Q_{t'}(h,a'),\ \frac{1}{1-\gamma}\Bigg\} - V_*(h)
    \geq -\frac{\tilde{\Delta}_\gamma}{1-\gamma}
$$ 
for all $h\in\histories$.  Therefore, the result holds for all $0\leq t\leq T$. \qedsymbol

Note that one direct implication of Lemma \ref{lemma:optimism} is that $\chi_t\geq 0$ for all $t=0,1,\dots,T$.

\subsubsection{A High-Probability Bound}
\label{subsec:high-probability-bound}
In this subsection we will prove the following lemma.

\begin{lemma}
\label{lemma:high-probability}
If Algorithm \ref{alg:fictional-q-learning-discounted} is executed with $\gamma$, $\beta_\delta$ and $Q_{\rm init}$ specified in Lemma \ref{lemma:optimism},
then with probability at least $1-\delta$,
$$
    \sum_{t=0}^{T-1} \big(\xi_t - \chi_t\big)
    \leq \frac{2-3\gamma}{1-\gamma}\cdot\tilde{\Delta}_\gamma T + \frac{\states\actions+1}{1-\gamma} + \frac{24}{(1-\gamma)^{3/2}}\cdot\sqrt{\states\actions T\cdot \log\frac{2T}{\delta}},
$$
where $\xi_t$ and $\chi_t$ are defined in \eqref{eq:def-xi} and \eqref{eq:def-chi}, respectively.
\end{lemma}
\noindentproof
For the moment let us fix $t\in\{0,1,\dots,T\}$, and consider the aleatoric state-action pair $\big( \phi(H_t), A_t\big)$, which has been updated $n \geq 1$ times before (and including) timestep $t$.  Let $1\leq t_1< \dots < t_n \leq t$ be the timestep in which $\big( \phi(H_t), A_t\big)$ is updated.  Recall that
\[
    \phi(H_{t_i}) = \phi(H_t),\quad A_{t_i} = A_t,\quad \forall i=1,\dots,n.    
\]
We have that, conditioned on event $\mathfrak{E}$, 
\[
    Q_t(H_t, A_t) - Q_*(H_t, A_t) 
    &=& \sum_{i=1}^{n} \alpha_{n}^i \cdot \left(
        R_{t_i+1} + \gamma\cdot V_{t_i}\big( H_{t_i+1}\big) + \frac{\beta_\delta}{\sqrt{i}} 
        - Q_*(H_t, A_t)
    \right)\nn\\
    &\leq& \left\{\sum_{i=1}^{n} \alpha_{n}^i \cdot \Big[
        \gamma\cdot V_{t_i}\big( h_{t_i+1}\big)
        - \gamma\cdot \trans V_*\big(H_{t_i}, A_{t_i} \big)
    \Big]\right\}\nn\\
        &&+\ \tilde{\Delta}_\gamma + \sum_{i=1}^{n} \alpha_{n}^i \cdot \frac{\beta_\delta}{\sqrt{i}} \label{eq:appendix-5}\\
    &\leq& \left\{\gamma\cdot\sum_{i=1}^{n}\alpha_{n}^i \cdot\Big[ V_{t_i}\big( H_{t_i+1}\big) - \trans  V_*\big(H_{t_i}, A_{t_i} \big) \Big]\right\}\nn\\
        &&+\ \tilde{\Delta}_\gamma + \frac{2\beta_\delta}{\sqrt{n}}\label{eq:appendix-6}\\
    &=& \left\{\gamma\cdot\sum_{i=1}^{n}\alpha_{n}^i \cdot\Big[ V_{t_i}\big( H_{t_i+1}\big) - V_*\big(H_{t_i+1}\big) \Big]\right\}+\tilde{\Delta}_\gamma + \frac{2\beta_\delta}{\sqrt{n}}\nn\\
        &&+\ \left\{\gamma\cdot\sum_{i=1}^{n}\alpha_{n}^i \cdot\Big[ V_*\big(H_{t_i+1}\big) - \trans V_*\big(H_{t_i}, A_{t_i} \big) \Big]\right\}\nn\\
    &\leq& \left\{\gamma\cdot\sum_{i=1}^{n}\alpha_{n}^i \cdot\Big[ V_{t_i}\big( H_{t_i+1}\big) - V_*\big(H_{t_i+1}\big) \Big] \right\}+ \tilde{\Delta}_\gamma + \frac{3\beta_\delta}{\sqrt{n}},\nn
    \\ \label{eq:convergence-1}
\]
where \eqref{eq:appendix-5} follows from the fact that
\[
    \phi(H_{t_i}) = \phi(H_t) \quad \Rightarrow\quad \left|Q_*(H_t,a) - Q_*(H_{t_i}, a) \right| \leq \Delta_\gamma\leq\tilde{\Delta}_\gamma,\ \forall a\in\actions;
\]
inequality \eqref{eq:appendix-6} follows from assertion $(a)$ of Lemma \ref{lemma:learning-rate}; and \eqref{eq:convergence-1} follows from that, conditioned on the event $\mathfrak{E}$, 
\[
    \left|\sum_{i=1}^{n}\alpha_{n}^i \cdot\Big[ V_*\big(H_{t_i+1}\big) - \trans V_*\big(H_{t_i}, A_{t_i} \big) \Big]\right|
    \leq \frac{\beta_\delta}{\sqrt{n}}.
\]
Combining \eqref{eq:appendix-13} and \eqref{eq:convergence-1}, we have
\[
    V_t(H_t) - V_*(H_t)
    &\leq& V_t(H_t) - Q_*(H_t, A_t)\nn\\
    &\leq& Q_t(H_t, A_t) - Q_*(H_t, A_t)\nn\\
    &\leq& \left\{\gamma\cdot\sum_{i=1}^{n}\alpha_{n}^i \cdot\Big[ V_{t_i}\big( H_{t_i+1}\big) - V_*\big(H_{t_i+1}\big) \Big]\right\} + \tilde{\Delta}_\gamma + \frac{3\beta_\delta}{\sqrt{n}}.
\]
For each $i=1,\dots,n$, if $\phi(H_{t_i+1}) = \phi(H_{t_i})$, then
\[
    \label{eq:appendix-19}
    V_{t_i}\big( H_{t_i+1}\big) - V_*\big(H_{t_i+1}\big) 
    &=& V_{t_i}\big( H_{t_i}\big) - V_*\big(H_{t_i+1}\big)\nn\\
    &\leq& V_{t_i}\big( H_{t_i}\big) - V_*\big(H_{t_i}\big) + \tilde{\Delta}_\gamma.
\]
Otherwise, if $\phi(H_{t_i+1}) \neq \phi(H_{t_i})$, then $\phi(H_{t_i+1})$ is not updated at timestep $t_i+1$, leading to 
\[
    \label{eq:appendix-20}
    V_{t_i}\big( H_{t_i+1}\big) = V_{t_i+1}\big(H_{t_i+1}\big). 
\]
Combining the two cases, we can claim that there exists $1\leq w_1<\dots < w_n\leq t+1$, such that
\[
    V_t(H_t) - V_*(H_t)
    &\leq& Q_t(H_t, A_t) - Q_*(H_t, A_t)\nn\\
    &\leq& \left\{\gamma\cdot\sum_{i=1}^{n}\alpha_{n}^i \cdot\Big[ V_{w_i}\big( H_{w_i}\big) - V_*\big(H_{w_i}\big) \Big]\right\} + 2 \tilde{\Delta}_\gamma + \frac{3\beta_\delta}{\sqrt{n}}.
    \label{eq:appendix-14}
\]
Note that \eqref{eq:appendix-14} only applies to times $t$ after which has been updated at least once.
Suppose that $\big(\phi(H_t), A_t\big)$ is first updated at time $t+1$ (meaning that it has not been updated even once prior to timestep $t$); then, naturally there should be
\[
    \label{eq:appendix-15}
    V_t(H_t) - V_*(H_t) \leq \frac{1}{1-\gamma}.
\]
Recall that $n$ is the number of times at which the value of $\big(\phi(H_t), A_t\big)$ is updated before (and including) timestep $t$.  We now let $t$ to take values in $0,1,\dots,T$, and replace $n$ and $w_i$ by $n_t$ and $w_{t,i}$ respectively to reflect their dependence on $t$. Summing all sides of \eqref{eq:appendix-14} from $t=0$ to $t=T-1$, and considering \eqref{eq:appendix-15}, we have
\[
    \sum_{t=0}^{T-1} \bigg(V_t(H_t) - V_*(H_t)\bigg)
    &\leq& \sum_{t=0}^{T-1} \bigg(Q_t(H_t, A_t) - Q_*(H_t, A_t)\bigg)\nn\\
    &\leq& \left\{\gamma\cdot\sum_{t=0}^{T-1} \sum_{i=1}^{n_t}\alpha_{n_t}^i\cdot \Big[ V_{w_{t,i}}\big( H_{w_{t,i}}\big) - V_*\big(H_{w_{t,i}}\big) \Big]\right\}\nn\\
        &&+\ 2\tilde{\Delta}_\gamma T + \frac{\states\actions}{1-\gamma} + \sum_{t=0}^{T-1}\frac{3\beta_\delta}{\sqrt{n_t}},
    \label{eq:appendix-16}
\]
where we note that there can be at most $\states\actions$ many ``first visits.''
We now want to determine whether $V_T(H_T) - V_*(H_T)$ appears in the summation on theright-hand side of \eqref{eq:appendix-16}.  Based on \eqref{eq:appendix-19} and \eqref{eq:appendix-20}, if it appears, there must be
\[
    w_{T-1, n_{T-1}} = T,
\]
meaning that the aleatoric state-action pair $\big(\phi(H_T), A_T\big)$ has never been visited in $t=0,1,\dots,T-1$, which leads to $V_T(H_T) - V_*(H_T)\leq 1/(1-\gamma)$.  Therefore, we can claim that
\[
    \sum_{t=0}^{T-1} \bigg(V_t(H_t) - V_*(H_t)\bigg)
    &\leq& \left\{\gamma\cdot\sum_{t=0}^{T-1} \sum_{i=1}^{n_t}\alpha_{n_t}^i\cdot \Big[ V_{w_{t,i}}\big( H_{w_{t,i}}\big) - V_*\big(H_{w_{t,i}}\big) \Big]\right\}\nn\\
        &&+\ 2\tilde{\Delta}_\gamma T + \frac{\states\actions+1}{1-\gamma} + \sum_{t=0}^{T-1}\frac{3\beta_\delta}{\sqrt{n_t}},
    \label{eq:appendix-21}
\]
where for all $t$ and $i$, $w_{t,i} \leq t-1$.

We have shown that conditioned on event $\mathfrak{E}$, $\chi_k\geq 0$ for all $k$.  Inequality \eqref{eq:appendix-16} implies that
\[
    \sum_{t=0}^{T-1} \xi_t
    &\leq& \left\{\gamma\cdot \sum_{t=0}^{T-1}\sum_{i=1}^{n_t} \alpha_{n_t}^i\cdot \left(\chi_{w_{t,i}} - \frac{\tilde{\Delta}_\gamma}{1-\gamma}\right) \right\} + 2\tilde{\Delta}_\gamma T + \frac{\states\actions+1}{1-\gamma} + \sum_{t=0}^{T-1}\frac{3\beta_\delta}{\sqrt{n_t}},\nn\\
    &=&\left\{\gamma\cdot \sum_{t=0}^{T-1}\sum_{i=1}^{n_t} \alpha_{n_t}^i\cdot \chi_{w_{t,i}}  \right\} + \frac{2-3\gamma}{1-\gamma}\cdot\tilde{\Delta}_\gamma T + \frac{\states\actions+1}{1-\gamma} + \sum_{t=0}^{T-1}\frac{3\beta_\delta}{\sqrt{n_t}},
    \label{eq:appendix-23}
\]
Examining the first term on the right-hand side of \eqref{eq:appendix-23}, from Lemma \ref{lemma:learning-rate} (c), there should be
\[
    \sum_{t=0}^{T-1}\sum_{i=1}^{n_t} \alpha_{n_t}^i\cdot \chi_{w_{t,i}} 
    \leq 
    \frac{3-\gamma}{2} \sum_{t=0}^{T-1} \chi_t.
    \label{eq:appendix-18}
\]
In terms of the last term in \eqref{eq:appendix-23}, letting $n_T(s,a)$ be the number of times aleatoric state-action pair $(s,a)$ is updated at times $t=1,\dots,T$, we have
\[
    \sum_{t=0}^{T-1}\frac{1}{\sqrt{n_t}} 
    &=& \sum_{s\in\states}\sum_{a\in\actions} \sum_{m=1}^{n_T(s,a)} \frac{1}{\sqrt{m}}\nn\\
    &\leq& \sum_{s\in\states}\sum_{a\in\actions} 2\sqrt{n_T(s,a)}\nn\\
    &\leq& 2\sqrt{\states\actions\cdot\sum_{s\in\states}\sum_{a\in\actions}n_T(s, a)}\nn\\
    &=& 2 \sqrt{\states\actions T},
    \label{eq:appendix-22}
\]
where in the final step we used the fact that
\[
    \sum_{s\in\states}\sum_{a\in\actions}n_T(s, a) = T. \nn
\]
Now we can revisit \eqref{eq:appendix-23}, starting from which there is
\[
    \sum_{t=0}^{T-1} \xi_t 
    &\leq& \left\{\gamma\cdot \sum_{t=0}^{T-1}\sum_{i=1}^{n_t} \alpha_{n_t}^i\cdot \chi_{w_{t,i}}  \right\} + \frac{2-3\gamma}{1-\gamma}\cdot\tilde{\Delta}_\gamma T + \frac{\states\actions+1}{1-\gamma} + \sum_{t=0}^{T-1}\frac{3\beta_\delta}{\sqrt{n_t}}\nn\\
    &\leq& \frac{\gamma(3-\gamma)}{2}\cdot \left(\sum_{t=0}^{T-1} \chi_t \right)+\frac{2-3\gamma}{1-\gamma}\cdot\tilde{\Delta}_\gamma T + \frac{\states\actions+1}{1-\gamma} + 6\beta_\delta\sqrt{\states\actions T}.\label{eq:appendix-24}
\]
Equivalently,
\[
    \sum_{t=0}^{T-1} \big(\xi_t - \chi_t\big)
    \leq \left(\frac{\gamma(3-\gamma)}{2}-1\right)\cdot \left(\sum_{t=0}^{T-1} \chi_t \right)+\frac{2-3\gamma}{1-\gamma}\cdot\tilde{\Delta}_\gamma T + \frac{\states\actions+1}{1-\gamma} + 6\beta_\delta\sqrt{\states\actions T}.\label{eq:appendix-27}
\]
Recall that conditioned on event $\mathfrak{E}$, $\chi_t\geq 0$ for all $t$, and further we have
\[
    \frac{\gamma(3-\gamma)}{2}-1 < 0,\quad \forall \gamma\in(0,1).\nn
\]
Thus, we can drop the first term in \eqref{eq:appendix-27} and deduce that, with probability $1-\delta$,
\[
    \sum_{t=0}^{T-1} \big(\xi_t - \chi_t\big)
    \leq \frac{2-3\gamma}{1-\gamma}\cdot\tilde{\Delta}_\gamma T + \frac{\states\actions+1}{1-\gamma} + \frac{24}{(1-\gamma)^{3/2}}\cdot\sqrt{\states\actions T\cdot \log\frac{2T}{\delta}}.
\]

\subsubsection{Finishing the Proof of Theorem \ref{theorem:discounted}}
It remains to show that the high-probability bound in Lemma \ref{lemma:high-probability} also implies a bound on the expected sum of $\xi_t - \chi_t$. For all $\delta>0$, let $\mathfrak{E}_\delta$ denote the event in \eqref{eq:high-probability-event}. Because
\[
    \xi_t - \chi_t = V_*(H_t) - Q_*(H_t, A_t) - \frac{\tilde{\Delta}_\gamma}{1-\gamma}\leq \frac{1}{1-\gamma},
\]
 we have that
\[
    \E\left[\left\{ \sum_{t=0}^{T-1} \big(\xi_t - \chi_t\big)\right\} \mathbf{1}(\mathfrak{E}^c) \right] 
    \leq \frac{T}{1-\gamma} \cdot \big( 1 - \Pr(\mathfrak{E}_\delta) \big)=\frac{\delta T}{1-\gamma}.
\]
Therefore,
\[
    \E\left[ \sum_{t=0}^{T-1} \big(\xi_t - \chi_t\big) \right]
    &=& \E\left[ \left\{\sum_{t=0}^{T-1} \big(\xi_t - \chi_t\big)\right\} \mathbf{1}( \mathfrak{E}) \right] + \E\left[\left\{ \sum_{t=0}^{T-1} \big(\xi_t - \chi_t\big)\right\} \mathbf{1}(\mathfrak{E}^c) \right] \nn\\
    &\leq& \frac{24}{(1-\gamma)^{3/2}}\cdot\sqrt{\states\actions T\cdot \log\frac{2T}{\delta}} + \frac{2-3\gamma}{1-\gamma}\cdot\tilde{\Delta}_\gamma T + \frac{\states\actions+1+\delta T}{1-\gamma}.\nn\\
\]
Letting $\delta = 1/T$, we have that
\[
    \beta = \frac{4}{(1-\gamma)^{3/2}}\sqrt{\log (2T^2)},
\]
and
\[
    \E\left[ \sum_{t=0}^{T-1} \big(\xi_t - \chi_t\big) \right] \leq \frac{24}{(1-\gamma)^{3/2}}\cdot\sqrt{\states\actions T\cdot \log(2T^2)} + \frac{2-3\gamma}{1-\gamma}\cdot\tilde{\Delta}_\gamma T + \frac{\states\actions+2}{1-\gamma}.
\]
Plugging the above inequality into \eqref{eq:after-decomposition-1}, we arrive at
\[
    \E_{\hat{\pi}} \left[ \sum_{t=0}^{T-1} \bigg(V_*(H_t) - V_{\hat{\pi}} (H_t)\bigg)\right] 
    &\leq& \frac{24}{(1-\gamma)^{\frac{5}{2}}}\cdot\sqrt{\states\actions T\cdot \log(2T^2)}\nn\\
        &&+\ \left(\frac{2-3\gamma}{(1-\gamma)^2} + \frac{1}{(1-\gamma)^2}\right)\cdot\tilde{\Delta}_\gamma T\nn\\
        &&+\ \frac{\states\actions+2}{(1-\gamma)^2}+ \frac{1}{(1-\gamma)^2}\nn\\
    &=& \frac{24}{(1-\gamma)^{\frac{5}{2}}}\cdot\sqrt{\states\actions T\cdot \log(2T^2)} \nn\\
        &&+\ \frac{3\tilde{\Delta}_\gamma T}{1-\gamma} + \frac{\states\actions+3}{(1-\gamma)^2},\nn
\]
which concludes the proof of Theorem \ref{theorem:discounted}. \qedsymbol

\subsection{From Discounted Return to Average Reward}
\label{subsec:discounted-to-average}
In this section we still continue to study the discounted $Q$-learning subroutine Algorithm \ref{alg:fictional-q-learning-discounted}, but we shift our focus to the learning performance with respect to the average reward.  Our goal is to show the following stronger version of Theorem \ref{theorem:average-reward-discounted-algorithm}.

\begin{theorem}
\label{theorem:average-reward-any-policy}
For all $\tau \geq 1$,
if Algorithm \ref{alg:fictional-q-learning-discounted} is executed with $\gamma = 1 - 1/\tau$,
\[
    \beta = 4\tau^{3/2}\sqrt{\log (2T^2)},
\]
and $Q_{\rm init}$ such that for some $\iota \geq 0$,
\[
    \label{eq:q-init-condition-2}
    Q_*^\gamma(h,a) - \iota\tau \leq Q_{\rm init}\big(\phi(h),a\big) \leq \tau,\quad \forall h\in\histories, a\in\actions,
\]
then for all $T>\tau\cdot\log(T)$, $\pi'\in\policies$ and initial history $h\in\histories$, we have that
\[
    \label{eq:theorem-discounted-stronger}
    \E_{\hat{\pi}} \left[ \sum_{t=0}^{T-1} \bigg(\lambda_{\pi'} - R_{t+1}\bigg)\right] 
    &\leq&
    24\tau^{3/2}\cdot\sqrt{\states\actions T\cdot \log(2T^2)} + 3\Big[\tilde{\Delta}_\tau + \tau_{\pi'}/\tau\Big]\cdot T \nn\\
    &&+\ \Big[\states\actions+5+2\log (T)\Big]\cdot\tau,
\]
where $\tilde{\Delta}_\tau = \max\{\Delta_\tau, \iota\}$ .
\end{theorem}

\noindentproof
First notice that for all $t\geq 0$,
\[
    \E_{\hat{\pi}}\Big[V_{\hat{\pi}}^\gamma(H_t)\Big] = \E\left[\sum_{\ell=0}^\infty \gamma^\ell R_{t + \ell + 1}\right].
\]
Thus, we have that
\begin{eqnarray}
    \E\left[\sum_{t=0}^{T-1} V_{\pi'}^\gamma(H_t) - V^\gamma_{\hat{\pi}}(H_t) \right]
    &\geq& \E\left[\sum_{t=0}^{T-1} \left(\frac{\lambda_{\pi'}}{1-\gamma} - \tau_{\pi'}\right)\right] - \E\left[\sum_{t=0}^{T-1}\sum_{\ell=0}^\infty \gamma^\ell R_{t+\ell+1}\right]\label{eq:appendix-10}\\
    &=& \E\left[\sum_{t=0}^{T-1} \sum_{\ell=0}^\infty \gamma^\ell \cdot \big(\lambda_{\pi'} - R_{t+\ell+1}\big) \right] - \tau_{\pi'}\cdot T\nn\\ 
    &=& \E\left[\sum_{t=0}^{T-1} \frac{1 - \gamma^{t+1}}{1 - \gamma} \cdot \big(\lambda_{\pi'} - R_{t+1}\big) \right]\nn\\
        &&+\ \E\left[\sum_{t=T}^{\infty} \gamma^{t+1-T}\frac{1 - \gamma^{T}}{1 - \gamma} \cdot \big(\lambda_{\pi'} - R_{t+1}\big) \right] \nn\\
        &&-\ \tau_{\pi'}\cdot T,
    \label{eq:compare-with-pi-1}
\end{eqnarray}
where \eqref{eq:appendix-10} is the result of Lemma \ref{lemma:mixing-time}.
Notice that, since $|\lambda_{\pi'} - R_t|\leq 1$,
\begin{eqnarray}
    \E\left[\sum_{t=T}^{\infty} \gamma^{t+1-T}\frac{1 - \gamma^{T}}{1 - \gamma} \cdot \big(\lambda_{\pi'} - R_{t+1}\big) \right]
    &\geq& -\sum_{t=T}^{\infty} \gamma^{t+1-T}\frac{1 - \gamma^{T}}{1 - \gamma}\nn\\
    &=& -\frac{1-\gamma^T}{1-\gamma}\cdot \frac{\gamma}{1-\gamma}\nn\\
    &\geq& -\frac{1}{(1-\gamma)^2}.
    \label{eq:compare-with-pi-2}
\end{eqnarray}
Let $T_0^\gamma = \lfloor\log(T)/(1-\gamma)\rfloor$, then for all $t > T_0^\gamma$, 
\[
    \gamma^{t} \leq \gamma^{\frac{\log T}{1-\gamma}} \leq \left(\frac{1}{e} \right)^{\log T}= \frac{1}{T}.
\]
Therefore,
\begin{eqnarray}
    \label{eq:compare-with-pi-3}
    \E\left[ 
        \sum_{t=T_0^\gamma}^{T-1} \frac{1 - \gamma^{t+1}}{1 - \gamma} \cdot \big(\lambda_{\pi'} - R_{t+1}\big)
    \right]
    &=& 
    \E\left[ 
        \sum_{t=T_0^\gamma}^{T-1} \frac{1}{1 - \gamma} \cdot \big(\lambda_{\pi'} - R_{t+1}\big)
    \right]\nn\\
        &&-\ \E\left[ 
        \sum_{t=T_0^\gamma}^{T-1} \frac{\gamma^{t+1}}{1 - \gamma} \cdot \big(\lambda_{\pi'} - R_{t+1}\big)
    \right]\nn\\
    &\geq& \frac{1}{1 - \gamma} \cdot \E\left[
        \sum_{t=T_0^\gamma}^{T-1}  \big(\lambda_{\pi'} - R_{t+1}\big)
    \right] - \sum_{t=T_0^\gamma}^{T-1} \frac{\gamma^{t+1}}{1 - \gamma}\nn\\
    &\geq& \frac{1}{1 - \gamma} \cdot \E\left[
        \sum_{t=T_0^\gamma}^{T-1}  \big(\lambda_{\pi'} - R_{t+1}\big)
    \right] - \frac{1}{(1 - \gamma)T}\cdot T\nn\\
    &\geq& \frac{1}{1 - \gamma} \cdot \E\left[
        \sum_{t=T_0^\gamma}^{T-1}  \big(\lambda_{\pi'} - R_{t+1}\big)
    \right] - \frac{1}{(1 - \gamma)}.
\end{eqnarray}
On the other hand,
\begin{equation}
    \label{eq:compare-with-pi-4}
    \E\left[ 
        \sum_{t=0}^{T_0^\gamma-1} \frac{1 - \gamma^{t+1}}{1 - \gamma} \cdot \big(\lambda_{\pi'} - R_{t+1}\big)
    \right] \geq -\frac{1}{1-\gamma}\cdot T_0^\gamma 
    \geq -\frac{\log(T)}{(1-\gamma)^2}.
\end{equation}
From Theorem \ref{theorem:discounted}, there is also
\begin{eqnarray}
    \label{eq:compare-with-pi-5}
    \E\left[\sum_{t=0}^{T-1} V_{\pi'}^\gamma(H_t) - V^\gamma_{\hat{\pi}}(H_t) \right]
    &\leq& 
    \E\left[ \sum_{t=0}^{T-1} V_*^\gamma(H_t) - V^\gamma_{\hat{\pi}}(H_t) \right]\nn\\
    &\leq& 24\tau^{5/2} \cdot\sqrt{\states\actions T\log (2T^2)} + 3\tilde{\Delta}_\tau\cdot\tau\cdot T + \Big[\states\actions+3\Big]\cdot\tau^2,\nn\\
\end{eqnarray}
where we note that $\tilde{\Delta}_\gamma = \tilde{\Delta}_\tau$ since $\gamma = 1 - 1/\tau$.
Combining \eqref{eq:compare-with-pi-1}-\eqref{eq:compare-with-pi-5}, we have that
\begin{eqnarray}
    \tau \cdot \E\left[
        \sum_{t=T_0^\gamma}^{T-1}  \big(\lambda_{\pi'} - R_{t+1}\big)
    \right] 
    &\leq& 24\tau^{5/2} \cdot\sqrt{\states\actions T\log (2T^2)} + 3\tilde{\Delta}_\tau\cdot\tau\cdot T + \Big[\states\actions+3\Big]\cdot\tau^2\nn\\
        &&+\Big[\log(T)+1\Big]\cdot\tau^2 + \tau + \tau_{\pi'}\cdot T.
\end{eqnarray}
For $T>\tau\cdot\log(T) \geq T_0^\gamma$,
\begin{eqnarray}
    \E\left[\sum_{t=0}^{T-1} \lambda_{\pi'} - R_{t+1} \right]
    &\leq& 24\tau^{3/2} \cdot\sqrt{\states\actions T\log (2T^2)} + 3\tilde{\Delta}_\tau\cdot T + \Big[\states\actions+3\Big]\cdot\tau\nn\\
        &&+\Big[\log(T)+1\Big]\cdot\tau + 1 + \tau_{\pi'}/\tau\cdot T + T_0^\gamma\nn\\
    &\leq& 24\tau^{3/2} \cdot\sqrt{\states\actions T\log (2T^2)} + 3\tilde{\Delta}_\tau\cdot T + \Big[\states\actions+5\Big]\cdot\tau\nn\\
        &&+2\log(T)\cdot\tau+ \tau_{\pi'}/\tau\cdot T,
\end{eqnarray}
which is what we claim in Theorem \ref{theorem:average-reward-any-policy}. \qedsymbol

\subsection{Proof of Lemma \ref{lemma:aleatoric-vs-global}}
The lemma is restated below.

\aleatoricvsglobal*

In fact, for a fixed $\tau \geq 1$, let $\gamma = 1 - 1/\tau < 1$.
Then, for all $\gamma'\in(\gamma,1)$ and $h_1, h_2\in\histories$, as long as $\phi(h_1) = \phi(h_2)$, there should be
\[
    \big|V_*^{\gamma'}(h_1) - V_*^{\gamma'}(h_1)\big| \leq \overline{\Delta}_\tau.
\]
From Theorem 7.1 in \citet{van2006performance}, for all $\epsilon>0$ and $\gamma'\in(\gamma, 1)$, there exists a policy $\tilde{\pi}_{\epsilon}^{\gamma'}\in\tilde{\policies}$ and a distribution $\mu_{\epsilon, \gamma'}$ on $\histories$, such that, for all $h\in\histories$,
\[
    \mu_{\epsilon, \gamma'}(h) > 0,
\]
and
\[
    (1-\gamma') \sum_{h\in\histories} \mu_{\epsilon, \gamma'}(h) \left[ V_*^{\gamma'}(h) - V_{\tilde{\pi}_{\epsilon}^{\gamma'}}^{\gamma'}(h) \right] \leq \gamma' \overline{\Delta}_\tau + \epsilon.
\]
Taking the limit $\gamma'\uparrow 1$ and noticing that, for all $\pi\in\policies$ and $h\in\histories$,
\[
    \lim_{\gamma'\uparrow 1} (1 - \gamma') V_\pi^{\gamma'}(h) = \lambda_\pi,
\]
we arrive at
\[
    \limsup_{\gamma'\uparrow 1} \lambda_{\pi_*^{\gamma'}}(h) - \lambda_{\tilde{\pi}_{\epsilon}^{\gamma'}}(h) \leq \overline{\Delta}_\tau + \epsilon.
\]
This means that, for all $\iota>0$, there exists $\gamma_\iota\in(\gamma,1)$, such that
\[
    \lambda_{\pi_*^{\gamma_\iota}}(h) - \lambda_{\tilde{\pi}_{\epsilon}^{\gamma_\iota}}(h) \leq \overline{\Delta}_\tau + \epsilon + \iota.
\]
Since $\lambda_{\tilde{\pi}_{\epsilon}^{\gamma_\iota}}\leq \lambda_{\tilde{\pi}}$, the above inequality implies that, for all $\epsilon, \iota > 0$,
\[
    \lambda_{\pi_*^{\gamma_\iota}}(h) - \lambda_{\tilde{\pi}}(h) \leq \overline{\Delta}_\tau + \epsilon + \iota.
\]
Now notice that, following the result in \citet{blackwell1962discrete}, when $\gamma_\iota \uparrow 1$, there should be $\lambda_{\pi_*^{\gamma_\iota}}(h) \to \lambda_{*}(h)$. Hence, we can take $\epsilon\downarrow 0$ and $\iota\downarrow 0$, which gives us
\[
    \lambda_{*}(h) -  \lambda_{\tilde{\pi}}(h) \leq \overline{\Delta}_\tau,
\]
as we desire. \qedsymbol

\subsection{Concluding the Proof of Theorem \ref{theorem:main}}
\label{subsec:proof-of-main-theorem}
In this section we complete the final steps towards Theorem \ref{theorem:main}, which we restate below.

\averagerewardtheorem*

With the set of subroutines ${\tt foo}_1$ to ${\tt foo}_4$ specified in \eqref{eq:foo-1}--\eqref{eq:foo-4}, the interactions between the agent and the environment can be viewed through epochs $k=1,2\dots$, with each epoch corresponding to executing Algorithm \ref{alg:fictional-q-learning-discounted} with a fixed a fixed discount factor $\gamma_k = 1-1/T_k^{1/5}$, which is equivalent to a fixed effective planning horizon $\tau_k = T_k^{1/5}$, where $T_k$ are the change points, with $T_0=1$ and $T_k = 20\times 2^{k-1}$, $k\geq 1$.  In the previous subsections we have analysed the average-reward regret of Algorithm \ref{alg:fictional-q-learning-discounted} when the discount factor is fixed.  In this subsection we will allow the discount factor to change and use a ``doubling trick'' to bound the total regret of $\pi_{\rm agent}$. Throughout this section, we assume that the total number of timesteps $T$ is fixed.  We start with the following useful lemma.

\begin{lemma}
    \label{lemma:doubling-trick}
    Let $a, S>0, b\geq 0$ be integers, $\zeta \in(0,1)$ and $f:\mathbb{Z}_+\mapsto \mathbb{R}$ is such that $f(x)\leq x^\zeta$ for all $x$.  Consider the sequence $t_j=a\cdot 2^{b+j}$, $j=0,1,\dots$ and let $k$ be the minimum index such that $t_0+t_1+\dots+t_k\geq S$.  If $k\geq 1$, then
    \[
        f(t_0) + \dots + f(t_k) \leq \frac{2^{2\zeta}}{2^{\zeta}-1}\cdot S^\zeta.
    \]
\end{lemma}
\noindentproof
First notice that 
\[
    t_0 + t_1 + \dots + t_k \leq 3S.
\]
This is because based on the definition, 
\[
    t_0 + t_1 + \dots + t_{k-1} = a\cdot2^b\cdot (2^k - 1)\geq a\cdot 2^{b+k-1} = \frac{1}{2} t_k.
\]
Hence, if $\sum_{t=0}^k t_k> 3S$, then $\sum_{t=0}^{k-1} t_k > S$, contradicting the minimality of $k$.  Since $\sum_{t=0}^k t_k = a\cdot 2^b\cdot (2^{k+1}-1)$, we have
\[
    k \leq 1 + \log_2 \frac{S}{a\cdot 2^b}.
\]
Therefore,
\begin{eqnarray}
    f(t_0) +\dots + f(t_k)
    \leq\sum_{t=0}^k t_k^{\zeta}
    \leq a^{\zeta}\cdot\frac{2^{(k+1)\zeta}}{2^{\zeta}-1} \leq \frac{2^{2\zeta}}{2^\zeta - 1}\cdot S^\zeta,
\end{eqnarray}
as desired. \qedsymbol

In light of our requirements on $Q_{\rm init}$ in Algorithm \ref{alg:fictional-q-learning-discounted}, we also need the following result.

\begin{lemma}
\label{lemma:q-init-condition}
For all $\gamma_1, \gamma_2$ such that $0\leq \gamma_1 < \gamma_2 < 1$, 
\[
    Q_*^{\gamma_1}(h,a)  - \frac{1}{1-\gamma_1} \geq Q_*^{\gamma_2}(h,a) - \frac{1}{1-\gamma_2}.
\]
\end{lemma}
\noindentproof
For $i=1,2$, let $\pi_i\in\policies$ be such that, for all $h\in\histories$, $\pi_i(\cdot|h)$ is the uniform distribution over $\argmax_{a'\in\actions} Q_*^{\gamma_i}(h,a')$.  From Proposition \ref{lemma:bellman}, $(\pi_i: i=1,2)$ exists, and
\[
    Q_*^{\gamma_i}(h,a) = \overline{r}_{h,a} + \sum_{h'\in\histories}P_{ahh'}\cdot\sum_{t=0}^\infty \gamma_i^{t+1} \big( P_{\pi_i}^t \overline{r}_{\pi_i} \big) (h'),\quad \forall i\in\{1,2\}, h\in\histories, a\in\actions.
\]
As a result, we have
\[
    Q_*^{\gamma_2}(h,a) - Q_*^{\gamma_1}(h,a)
    &\leq& Q_{\pi_2}^{\gamma_2}(h,a) - Q_{\pi_2}^{\gamma_1}(h,a)\nn\\
    &=& \sum_{h'\in\histories}P_{ahh'}\cdot\sum_{t=0}^\infty \Big\{ \big(\gamma_2^{t+1} - \gamma_1^{t+1} \big)\cdot \big( P_{\pi_2}^t \overline{r}_{\pi_2} \big) (h')\Big\}\nn\\
    &\leq& \sum_{h'\in\histories}P_{ahh'}\cdot\sum_{t=0}^\infty \Big\{ \big(\gamma_2^{t+1} - \gamma_1^{t+1} \big)\cdot 1\Big\}\nn\\
    &=& \frac{\gamma_2}{1-\gamma_2} - \frac{\gamma_1}{1-\gamma_1}\nn\\
    &=& \frac{1}{1-\gamma_2} - \frac{1}{1-\gamma_1},
\]
which is what we desire. \qedsymbol

The above lemma, together with  Lemma \ref{lemma:optimism} and the subroutine ${\tt foo}_3$ defined in \eqref{eq:foo-3}, ensures that at the beginning of epoch $k$,
\[
    Q\big(\phi(h),a\big) \geq Q_*^{\gamma_k}(h,a) - \overline{\Delta}_{\tau_k}\cdot\tau_k,\quad \forall k=0,1,\dots, h\in\histories, a\in\actions.
\]

Let $\mathcal{T}_k$ denote the timesteps in the $k$-th epoch, $k=0,1,\dots, L$, where $L$ is the index of the epoch that contains timestep $T-1$ (which is the last epoch that we are concerned about), i.e.
\[
    \mathcal{T}_0 &=& \{0,1,\dots, 19\},\nn\\
    \mathcal{T}_1 &=& \{20,21,\dots, 39\},\nn\\
    \mathcal{T}_2 &=& \{40,41,\dots, 79\},\nn\\
        &\cdots&\nn\\
    \mathcal{T}_L &=& \{20\cdot 2^{L-1}, 20\cdot 2^{L-1}+1, \dots, T-1\}.\nn
\]
Letting $|\mathcal{T}|$ be the length of epoch $\mathcal{T}$, we have that $|\mathcal{T}_k| = T_k$ for $1\leq k\leq L-1$ and $|\mathcal{T}_0|=20$, $|\mathcal{T}_L|=T - 20\cdot 2^{L-1}$.  For a fixed reference policy $\pi\in\policies$, let
\[
    \mathcal{R}_{\pi}(\mathcal{T}_k) = \E\left[\sum_{t\in\mathcal{T}_k} \big(\lambda_{\pi} - R_{t+1}\big)\Big|\environment\right].
\]
For $k\geq 1$, let 
\[
    \gamma_k = 1 - \frac{1}{T_{k-1}^{1/5}} = 1 - \frac{1}{|\mathcal{T}_k|^{1/5}}
\]
be the discount factor used in the $k$-th epoch (where $\gamma_0=0$ is the discount factor in the $0$-th epoch).  By Theorem \ref{theorem:average-reward-any-policy}, either
\begin{eqnarray}
    \mathcal{R}_\pi(\mathcal{T}_k) 
    &\leq& 24\tau_k^{3/2} \cdot\sqrt{\states\actions|\mathcal{T}_k|\log (2|\mathcal{T}_k|^2)} + \Big[3\overline{\Delta}_{\tau_\pi} + \tau_{\pi}/\tau_k\Big]\cdot |\mathcal{T}_k| \nn\\
        &&+ \Big[\states\actions + 2\log(|\mathcal{T}_k|)+5\Big]\cdot\tau_k \nn\\
    &\leq& \Big(24 \cdot\sqrt{\states\actions\log (2T^2)} + \tau_{\pi} \Big)\cdot |\mathcal{T}_k|^{4/5}+ 3\overline{\Delta}_{\tau_\pi}\cdot |\mathcal{T}_k| \nn\\
        &&+ \Big[6\states\actions + 2\log(T)\Big]\cdot |\mathcal{T}_k|^{\frac{1}{5}},
    \label{eq:one-epoch-regret}
\end{eqnarray}
or, if $\tau_k \leq \tau_\pi$,
\[
    \mathcal{R}_\pi(\mathcal{T}_k) \leq |\mathcal{T}_k|.
\]
Let 
\[
    \mathbb{I}_k = 
    \begin{cases}
        1 & \text{if $\tau_k > \tau_\pi$}\\
        0 & \text{otherwise}
    \end{cases},
\]
and let
\[
    g(t) = \Big(24 \cdot\sqrt{\states\actions\log (2T^2)} + \tau_{\tilde{\pi}_*} \Big)\cdot t^{\frac{4}{5}}+ 3\Delta\cdot t + \Big(6\states\actions + 2\log(T)\Big)\cdot t^{\frac{1}{5}}.
\]
We arrive at
\begin{eqnarray}
    \regret_\pi(T)
    &\leq& \sum_{k=0}^L \mathcal{R}_\pi(\mathcal{T}_k)\nn\\
    &\leq& \sum_{k=0}^L|\mathcal{T}_k|\cdot \mathbf{1}(\mathbb{I}_k=0) + g(|\mathcal{T}_k|)\cdot \mathbf{1}(\mathbb{I}_k=1) \nn\\
    &=& \sum_{k=0}^L|\mathcal{T}_k|\cdot \mathbf{1}\left\{\tau_k \leq \tau_{\pi}\right\} + g(|\mathcal{T}_k|)\cdot \mathbf{1}\left\{\tau_k > \tau_{\pi}\right\} \nn\\
    &=& \sum_{k=0}^L|\mathcal{T}_k|\cdot \mathbf{1}\left\{|\mathcal{T}_k| \leq \tau_{\pi}^5\right\} + g(|\mathcal{T}_k|)\cdot \mathbf{1}\left\{|\mathcal{T}_k| > \tau_{\pi}^5\right\}\nn\\
    &\leq& 2\tau_{\pi}^5 + \sum_{k=0}^L g(|\mathcal{T}_k|)\cdot \mathbf{1}\left\{|\mathcal{T}_k| > \tau_{\pi}^5\right\}.
\end{eqnarray}
By Lemma \ref{lemma:doubling-trick},
\[
    \sum_{k=0}^L g(|\mathcal{T}_k|)\cdot \mathbf{1}\left\{|\mathcal{T}_k| > \tau_{\pi}^5\right\}
    \leq \Big(120 \sqrt{\states\actions\log (2T^2)} + 5\tau_{\pi} \Big)\cdot T^{\frac{4}{5}}+ 3\Delta T + \Big(54\states\actions + 18\log(T)\Big)\cdot T^{\frac{1}{5}}.\nn
\]
Therefore, we have 
\begin{eqnarray}
    \regret_\pi(T)
    &\leq&  \Big(120 \sqrt{\states\actions\log (2T^2)} + 5\tau_{\pi} \Big)\cdot T^{\frac{4}{5}} + \Big(54\states\actions + 18\log(T)\Big)\cdot T^{\frac{1}{5}} \nn\\
        &&+ 
        3\Delta\cdot T + 2\tau_{\pi}^5,
\end{eqnarray}
which justifies our claim in Theorem \ref{theorem:main}. \qedsymbol

\section{Results on Approximate Dynamic Programming}
\label{sec:adp}

\begin{theorem}
For all $N\geq 2$, there exists an environment $\environment=(\actions,\observations,\rho)$, a set of aleatoric states $\states$, an aleatoric state update function $f$ and a reward function $r$, such that for all $\tau\geq 1$,
\[
    \lambda_* - \lambda_{\tilde{\pi}} =  \overline{\Delta}_\tau.
\]
\end{theorem}
{\bf Proof.}\ 
Consider an environment with two actions $\actions=\{1,2\}$ and two observations $\observations=\{0,1\}$.  The observation probabilities are given by
\[
    \rho(1|h,a)
    =
    \begin{cases}
    1 & \text{if $a_{|h|-1}(h) \neq a$}\\
    0 & \text{otherwise}
    \end{cases},\quad \forall h\neq\emptyH,
\]
and
\[
    \rho(0|\emptyH,a) = 1.
\]
In other words, the observation is deterministically 1 if and only if the agent takes a different action from the one that it took in the previous timestep, and the observation is always 0 in the first timestep.  There is only one aleatoric state $\states=\{1\}$, and the aleatoric state update function is
\[
    f(s,a,o)=1,\quad \forall s,a,o.
\]
The reward is equal to the observation, i.e.
\[
    r(s,a,o) = o.
\]
Notice that $\lambda_* = 1$, which can be attained by alternating between the two actions in each timestep.

We claim that $\Delta_\gamma = 1$ for all $\gamma\in(0,1)$.  Since there is only one aleatoric state, we only have to verify that, for all history pairs $(h_1, h_2)$ and action $a\in\{1,2\}$, 
\[
    \label{eq:verify-delta}
    \big|Q_*^\gamma(h_1, a) - Q_*^\gamma(h_2, a)\big| \leq 1.
\]
Indeed, for all history $h$ and action $a$, we have
\[
    Q_*^\gamma(h,a) = 
    \begin{cases}
    \frac{1}{1-\gamma} & \text{if $h\neq \emptyH$ and $a \neq a_{|h|-1}(a)$}\\
    \frac{\gamma}{1-\gamma} & \text{otherwise}
    \end{cases}.
\]
Since $Q_*^\gamma(h,a)$ can only take the above two values, \eqref{eq:verify-delta} obviously holds.  Thus,
\[
    \overline{\Delta}_\tau = 1,\quad \forall \tau\geq 1.
\]
However, there are only two policies in $\policies_{\rm aleatoric}$, one always taking action 1, and the other always taking action 2.  Either policy results in an all-zero reward sequence, implying that $\lambda_{\tilde{\pi}} = 0$.  Thus,
\[
    \lambda_* - \lambda_{\tilde{\pi}} = 1,
\]
as we desire.\qedsymbol

Next we show that an intuitive approximate dynamic programming algorithm may generate a policy whose performance is much worse than $\tilde{\pi}$, even in the simpler case where the environment dynamics are known. The example is adapted from the one in \cite{van2006performance}.

Specifically, we consider an MDP with $2N$ states $\{1,2,\dots, 2N\}$ and two actions $\{1,2\}$, where $N\geq 2$ is an integer. 
Note that this MDP can be viewed as an environment $\environment=(\actions,\observations,\rho)$, where $\actions=\{1,2\}$, $\observations = \{1,2,\dots,2N\}$ with observations corresponding to the MDP states (which we will simply call ``states'' hereafter), and $\rho$ depends on the history only through the most recent observation.
The $2N$ observations (or states) induce $2N$ equivalence classes in the set of histories $\histories$, where $h\sim h'$ if and only if their last observations are the same.
Henceforth we will use observation $o$ to denote the equivalence class in $\histories$ induced by $o$.  We will also not distinguish between ``observations'' and ``states.''

In every timestep, with probability $\epsilon_1$ the system ``resets'' itself, and the next state is drawn uniformly from $\{1,2,\dots,2N\}$.  Conditioned on that the system does not reset, from states $3,5,\dots,2N-1$ the system transitions deterministically to state 1, and from states $4,6,\dots,2N$ the system transitions deterministically to state 2, regardless of the action; from state 1 the system transitions to state 2 with probability $\epsilon_2$ under both actions; and from state 2, the system transitions to state 1 deterministically under action 1, and stays in state 2 deterministically under action 2.  The environment dynamics $\rho$ can thus be written as
\[
    \rho(1|o,a) = 
    \begin{cases}
    (1-\epsilon_1)(1-\epsilon_2) + \frac{\epsilon_1}{2N} & \text{if $o = 1$}\\
    1-\epsilon_1 + \frac{\epsilon_1}{2N} & \text{if $\big(o \in \{3,5,\dots, 2N-1\}\big)$ or $\big(o=2, a = 1\big)$}\\
    0 & \text{otherwise}
    \end{cases},\nn
\]
\[
    \rho(2|o,a) = 
    \begin{cases}
    (1-\epsilon_1)\epsilon_2 + \frac{\epsilon_1}{2N} & \text{if $o = 1$}\\
    1-\epsilon_1 + \frac{\epsilon_1}{2N} & \text{if $\big(o \in \{4,6,\dots, 2N\}\big)$ or $\big(o=2, a = 2\big)$}\\
    0 & \text{otherwise}
    \end{cases},\nn
\]
and
\[
    \rho(x|o,a) = \frac{\epsilon_1}{2N},\quad \forall x\in\{3,4,5,\dots,2N\}, o\in\observations, a\in\actions.\nn
\]
Let $\delta, \kappa>0$.  From every state in $\{4,6,\dots,2N\}$, the agent receives reward $\delta$ if it takes action 1, and reward $-\kappa$ if it takes action 2. From every state in $\{3,5,\dots,2N-1\}$, the agent receives reward $-\delta$ regardless of the action.  Additionally, the agent also receives reward $-\kappa$ if it takes action 2 in state 2.  In all other scenarios the agent receives zero reward.  The aleatoric state space is $\states=\{1,2\}$, with $\phi(o) = 1$ whenever $o\in\{1,3,\dots,2N-1\}$ and $\phi(o) = 2$ whenever $o\in\{2,4,\dots,2N\}$.

The learning algorithm that we consider proceeds as follows.  In iteration $k$, the algorithm independently samples $o_1^{(k)},\dots, o_M^{(k)}\in\observations$ according to the uniform distribution.  The value function is then updated via
\[
    V^{(k)} \leftarrow \argmin_{V:\states\mapsto\mathbb{R}} \sum_{m=1}^M \left[V\big(\phi(o_m^{(k)})\big) - \max_{a\in\actions} \left\{\overline{r}_{ao_m^{(k)}} + \gamma\cdot \trans \big(V^{(k-1)}\circ \phi\big)(o_m^{(k)}, a)\right\}\right]^2,
\]
where $\gamma\in(0,1)$ is a fixed discount factor.
Note that the algorithm is able to compute $\overline{r}$ and $\trans$ since the environment dynamics function $\rho$ is known.  This algorithm is a version of the fitted value iteration algorithm with $\ell_2$-norm loss function, as is studied in \citep{munos2008finite}.
It is worth mentioning that the sequence $\big(V^{(k)}:k=0,1,2,\dots\big)$ need not converge for all initial value functions $V^{(0)}$.
However, if $\big(V^{(k)}:k=0,1,2,\dots\big)$ converges in probability to $V^{(\infty)}:\states\to\mathbb{R}$, we have the following result.

\begin{theorem}
\label{theorem:adp-fail}
For all $\gamma\in(0,1)$ and $\epsilon>0$, there exists integers $M, N$ such that, for all $\tau\geq 1$ and $\pi^{(\infty)}$ greedy with respect to $V^{(\infty)}$,
\[
    \lambda_* - \lambda_{\pi^{(\infty)}} \geq \tau_{\pi^{(\infty)}}\cdot\gamma\cdot \overline{\Delta}_\tau - \epsilon.
\]

\end{theorem}

\noindentproof
Fix $\gamma\in[0,1)$.
We can verify that
\[
    Q_*^\gamma(1, a) = 0,\quad a\in\{1,2\},\nn
\]
and
\[
    Q_*^\gamma(k, a) = -\delta,\quad\forall k\in\{3,5,\dots, 2N-1\},  a\in\{1,2\}.\nn
\]
There is also
\[
    Q_*^\gamma(2, 1) = 0,\quad Q_*^\gamma(2, 2) = -\kappa,\nn
\]
and
\[
    Q_*^\gamma(k, 1) = \delta,\quad
    Q_*^\gamma(k, 2) = -\kappa,\quad
    \forall k\in\{4,6,\dots,2N\}.\nn
\]
Since all states in $\{1,3,\dots,2N-1\}$ are mapped to aleatoric state 1 and all states in $\{2,4,\dots,2N\}$ are mapped to aleatoric state 2, we have that
\[
    \overline{\Delta}_\tau = \delta,\nn
\]
for all $\tau\geq 1$.

The optimal average reward in this environment is $\lambda_* = 0$, which is attained by applying action 1 in every state.  We can consider $\pi'\in\tilde{\policies}$, which chooses action 1 in aleatoric state 1 and action 2 in aleatoric state 2.  Apparently $\lambda_{\pi'} < -\kappa/2$. In addition, \cite{van2006performance} established that whenever $\kappa < 2\gamma\delta / (1-\gamma)$ and $\epsilon_1 = 1-\gamma$, there exists $N$ and $M$ such that $\pi^{(\infty)}= \pi'$.  Since in this environment $\tau_{\pi'} = 1/\epsilon_1 = 1/(1-\gamma)$, we arrive at our desired result. \qedsymbol

\vskip 0.2in
\bibliography{references}

\end{document}